\documentclass[journal,twoside, print]{ieeecolor}
\usepackage{generic}
\usepackage{cite}
\usepackage{amsmath,amssymb,amsfonts}
\usepackage{algorithm}
\usepackage{algpseudocode}
\usepackage{graphicx}
\usepackage{subcaption}
\usepackage{multirow}
\usepackage{mathrsfs}
\usepackage{textcomp}
\usepackage{booktabs}
\usepackage{hyperref}
\usepackage[table]{xcolor}
\usepackage{colortbl}
\usepackage{lmodern}
\usepackage{placeins}
\usepackage{titlesec}
\usepackage{listings}
\usepackage{manyfoot}

\def\BibTeX{{\rm B\kern-.05em{\sc i\kern-.025em b}\kern-.08em
    T\kern-.1667em\lower.7ex\hbox{E}\kern-.125emX}}

\begin{document}

\title{EEG Artifact Detection and Correction with Deep Autoencoders}

\author{David Aquilué-Llorens, Aureli Soria-Frisch
\thanks{David Aquilué-Llorens and Aureli Soria-Frisch are with Starlab Barcelona S.L. Neuroscience BU, Av. Tibidabo 47 bis, Barcelona, 08035, Spain. (e-mail: aureli.soria-frisch@starlab.es)}}

\maketitle

\begin{abstract}
EEG signals convey important information about brain activity both in healthy and pathological conditions. However, they are inherently noisy, which poses significant challenges for accurate analysis and interpretation. Traditional EEG artifact removal methods, while effective, often require extensive expert intervention. This study presents LSTEEG, a novel LSTM-based autoencoder designed for the detection and correction of artifacts in EEG signals. Leveraging deep learning, particularly LSTM layers, LSTEEG captures non-linear dependencies in sequential EEG data. LSTEEG demonstrates superior performance in both artifact detection and correction tasks compared to other state-of-the-art convolutional autoencoders. Our methodology enhances the interpretability and utility of the autoencoder's latent space, enabling data-driven automated artefact removal in EEG its application in downstream tasks. This research advances the field of efficient and accurate multi-channel EEG preprocessing, and promotes the implementation and usage of automated EEG analysis pipelines for brain health applications.
\end{abstract}


\section{Introduction}
\label{sec:introduction}
\IEEEPARstart{E}lectroencephalography (EEG) is a non-invasive and cost-effective approach to register the electrical activity generated in the brain. This activity is highly informative and is leveraged across a wide range of applications such as diagnosing brain diseases \cite{Acharya2015, Meghdadi2021},  monitoring epilepsy \cite{Noachtar2009,Maganti2013, Acharya2013}, determining quality of sleep \cite{Campbell2009, Phan2022}, affective computing \cite{Liu2021, Gannouni2021}, and interacting with machines through Brain-Computer Interfaces \cite{Lazarou2018}, among others. However, the electrical potentials generated by neural activity present low amplitudes and are usually overshadowed by artifacts: spurious sources of high-amplitude electrical noise originated by  ocular or muscular activity, electromagnetic interferences, incorrect EEG electrode contact, etc \cite{Jiang2019}. These artifacts hinder the adequate analysis of the brain recordings and should, therefore, be removed as accurately as possible, which is attained through the offline application of different methodologies. The occurrence of artifacts is greatly exacerbated in EEG studies in more naturalistic scenarios, which have been gaining popularity thanks to advancements in cost and portability of recording devices \cite{Sawangjai2020, Dadebayev2022, Stangl2023}.
Therefore, artifact removal is crucial to spread the usage of accessible, non-invasive brain monitoring in new application domains. Currently, EEG artifact removal is a time-consuming process requiring expert evaluation. Developing tools to automate this task can significantly alleviate the burden on researchers and clinical practitioners, facilitating automated EEG analysis. Specially in out-of-the-lab settings the performance of artefact correction methodologies is not always optimal and offer clear room for improvement. Furthermore, the implementation of real-time artefact removal paves the path to effectively use EEG in ergonomics, human-machine interaction, and clinical decision support systems, to name a few.

Classical artifact removal techniques rely on identifying specific sources of noise to apply targeted strategies. Techniques such as filtering are effective against power line noise, which appear at very specific frequencies. However, removing other artifactual subsignals originated by e.g. eye blinks or muscle activation, requires more sophisticated methods. Among diverse Blind Source Separation methodologies \cite{Fitzgibbon2007}, Independent Component Analysis (ICA) has been widely used as a reliable statistical algorithm to decompose an EEG recording into its multiple independent constituents \cite{Hyvrinen2000, Jiang2019}. This approach allows to remove recurring artifacts, such as eye blinks \cite{Makeig_1995}, but requires expert knowledge on artefact characterization, careful pre-processing, significant computational resources for long recordings or large datasets, and does not allow the implementation of automated pipelines.

In recent years, multiple tools leveraging current advancements on deep learning have been developed to automate EEG artifact removal. Pion-Tonachini et al. \cite{PionTonachini2019} introduced ICLabel, a Convolutional Neural Network (CNN) designed to complement ICA by automatically classifying source signals. Despite its successful implementation to automate the pre-processing of large datasets, the method is still constrained by ICA's limitations, particularly its reliance on simple mapping functions between sources and channel recordings that may simplify the complex and non-linear nature of neuronal activity \cite{Chuang2022}.

Thus, recent work has focused on developing deep neural networks, capable of learning highly complex non-linear relations to automatically remove artifacts from EEG signals. Initial efforts targeted single-channel EEG corrections, exemplified by Zhang and colleagues' work, introducing the EEGDenoiseNet benchmark dataset for artifact removal \cite{Zhang2021} and using it to train multiple networks for automated EEG segment correction. Following studies leveraged the EEGDenoiseNet dataset to train neural networks able to separate the contaminated from the neurophysiological signal \cite{Yu2022} and Generative Adversarial Networks (GANs) that learn to generate a clean epoch from a noisy input to deceive an adversarial discriminator network \cite{Brophy2022}.

However, this single-channel focus overlooks crucial information in the spatial domain, essential for accurately correcting artifacts across channels placed at different points on the scalp. Despite the higher computational demands and complexity, a multi-channel approach promises a more holistic and effective strategy for EEG artifact removal.

Several works have implemented Convolutional Neural Networks (CNNs) for multi-channel artifact correction. Saba Sadiya et al. \cite{SabaSadiya2021} applied an ensemble of CNNs to correct artifacts in EEG epochs after an artifact detection step, while Lopes and colleagues developed a Deep CNN to remove artifacts from long epilepsy recordings \cite{Lopes2021}. Nonetheless, Lopes et al.'s work highlights the inability of the network at removing large artifacts, indicating a need for an additional rejection step to remove segments where brain activity is totally masked by artifacts.

More recent research has been devoted to autoencoders, a particular type of deep neural network architecture. Its ability to compress and decompress data efficiently by learning a low-dimensional representation of input features in an unsupervised manner eases the implementation of automated correction algorithms circumventing the effort-intensive process of manual data labelling.
The work in \cite{Chuang2022} presents IC-U-Net, an adaptation of the widely successful UNET architecture \cite{Ronneberger2015} for automatic multi-channel EEG artifact correction. By training the UNET with a dataset of raw and denoised signal pairs, created using ICLabel, the network learns to denoise EEG signals by minimizing the Mean Squared Error (MSE) between its outputs and the target denoised signals. Similarly, Lai and colleagues successfully implemented a lightweight autoencoder CNN \cite{Lai2022} demonstrating comparable performance to IC-U-Net using a significantly smaller network architecture.

In this present study, we further advance the field of automated multi-channel EEG artifact removal through the development of  \textbf{LSTEEG} (Starlab's Artifact Removal Autoencoder for EEG recordings), a novel LSTM-based Autoencoder designed for detection and correction of artifacual activity in EEG signals.
Our novel implementation leverages LSTM layers, which are specifically designed to capture long-term non-linear dependencies in sequential data \cite{Hochreiter1997, VanHoudt2020} and have already demonstrated EEG processing capabilities for classification \cite{Kumar2019, Chakravarthi2022} and denoising \cite{Zhang2021}. We supersede this last work by presenting the first LSTM autoencoder to work with multi-channel EEG, which incorporates the aforementioned advantages.

LSTEEG learns to encode each EEG segment into a homogeneous low-dimensional latent space, thereby harnessing the benefits of AEs, such as improved interpretability, thanks to its lower-dimensional representations, and synthetic EEG sample generation, both clean and artifactual, which can be further exploited for data augmentation.
Moreover, we propose a novel method to leverage the unsupervised capabilities of AEs in the context of artifact removal. Recognizing the importance of identifying instances of incomplete denoising, we incorporate an anomaly detection technique to detect EEG segments containing artifactual information, serving as a crucial step either before or after the correction process to guarantee thorough pre-processing.

We conduct a comparative performance evaluation of \textbf{LSTEEG} with recently developed state of the art convolutional AEs in two different EEG data sets. We demonstrate that our LSTEEG presents competitive performances in both artifact detection and correction tasks, while also learning meaningful low dimensional representations in its Latent Space. These representations add an interpretability layer for further understanding of what constitutes clean EEG signals, as well as to further extract data-driven neurophysiological features, which can be later used in downstream tasks.   

Our contribution is three-fold:\\
- We present LSTEEG, a novel LSTM-based autoencoder network for EEG artifact detection and correction with a clearly defined compressed Latent Space.\\
- We show that unsupervised training of autoencoders with clean EEG epochs, which does not require the usage of ground truth labels, allows for accurate classification of contaminated signals.\\
- We propose multiple ways to study and interpret the learned Latent Space of our LSTM-based autoencoder, characterizing clean EEG with a data-driven approach while opening up novel ways to utilize the learned features for its further application.

\section{Methods}\label{sec:Methods}

\subsection{Artifact Detection}\label{sec:ArtifactDetection}
Anomaly Detection differs from the classical supervised classification approaches where examples of the different classes are given during the training procedure. Anomaly Detection approaches learn to characterize the "normal" class, from which it detects deviations during the inference phase. It includes data-driven algorithms that detect atypical patterns in observations, which are likely caused by a different mechanism \cite{Pang2021}. Thus, we propose a novel approach to detecting EEG artifacts as an anomaly detection problem. Since labeled data in the context of artifactual EEG is expensive to generate, we propose an unsupervised approach based on autoencoders. Autoencoders are trained to minimize the reconstruction error between the outputs they produce and the input they have been fed. Thus, when an input has characteristics that significantly differ from the characteristics of the samples in their training data, the reconstruction is degraded. It is possible then to leverage the reconstruction error value as an anomaly metric, i.e used to classify the input as being an anomalous or regular observation through binary classification.

Therefore, we train the implemented AEs with already pre-processed clean EEG data from the LEMON dataset, fully described in section \ref{sec:LemonDataset}. A 60/20/20 training/validation/test hold-out partition is applied to the clean LEMON dataset for performance evaluation. We use 60\% for training, 20\% as a validation set for early stopping, and 20\% for testing. During training, we aim to find the networks' weights that minimize the Mean Squared Error (MSE) between the AE's input and output epochs. Additionally, we merge the EEG epochs in the clean testing dataset with noisy epochs from the original LEMON data, to create the  "LEMON \textit{Clean/RawFiltered}" dataset, which we use to validate the artifact detection capabilities of the implemented networks. 

Details on the training process and hyperparameters are provided in the Supplementary Information, Supp1.1.

After training, we forward the evaluation epochs through the AEs. AEs are expected to produce accurate reconstructions for EEG epochs with similar characteristics to the training data (low reconstruction MSE). However, the networks will provide degraded reconstruction (high reconstruction MSE) for epochs containing artifacts, as these anomalies were not present in the training process. Thus, the reconstruction MSE becomes the predictive function used to classify epochs: lower MSE values indicate a higher probability of an epoch being clean while higher MSE values indicate a higher probability of being noisy. The Area Under the Receiver Operating Characteristic Curve (AUC) will be used to determine the predictive ability of the trained network.

\subsection{Artifact Correction}
In the context of artifact correction, our objective is to train the deep networks to eliminate artifacts from EEG signals while preserving the intrinsic brain activity. To achieve this task, we must take a supervised training approach, where each EEG epoch serves as input, and the network learns to approximate its output to a predetermined target, specifically the artifact-free version of the input epoch.

We generate this dataset by automatically denoising the LEMON dataset using ICLabel, following similar steps as in \cite{Chuang2022}, thus creating an extensive collection of epoch pairs: the original, uncleaned input epoch and its corresponding, automatically denoised target epoch. It is worth mentioning that this dataset also includes epochs originally unaffected by noise, untouched by the automatic denoising process. These epochs are invaluable, as they provide the network with examples of clean signals, demonstrating cases where no corrective action is required.

Following the pipeline used in the Artifact Detection problem, we partition the epoch pairs into training, validation and testing sets in a 60\%/20\%/20\% partition. After training, we assess the performance of the models using the average root mean square error (RMSE) across the test set epochs. We quantitatively compare the error scores between the different implemented model to determine the best performing approaches. We will further examine the reconstruction capacity of the models by qualitatively analyzing the time and frequency domain visualizations of the reconstructed epochs.

\subsection{Datasets}\label{sec:Datasets}

\subsubsection{LEMON dataset}\label{sec:LemonDataset}
The LEMON dataset is a publicly available dataset that contains multiple physiological recordings of 227 healthy subjects, as well as the results of a battery of cognitive and psychological tests \cite{Babayan2019}. The resting state EEG has a duration of 16 minutes, using a 61-channel montage in a 10-10 electrode system. The recording is done with interleaved eyes-open and eyes-closed blocks of 60s during a resting-state task. The  dataset includes both raw and pre-processed EEG signals.

\paragraph{Pre-Processed LEMON for Artifact Detection}
In order to train a network for artifact detection through anomaly detection, we will need a training data set as clean as possible. Thus, we use the pre-processed data that has been visually inspected and manually cleaned by Babayan et al.

In \cite{Babayan2019}, the pre-processing pipeline applied consisted in downsampling the signals from 2500Hz to 250Hz and applying a bandpass filter between 1-45Hz (using an 8th order Butterworth filter). Outlier channels were rejected after visual inspection for frequent jumps/shifts in voltage and poor signal quality. Additionally, data intervals with extreme peak to peak deflections or large bursts of high frequency activity were also identified by visual inspection and removed. Finally, PCA and ICA were applied to remove components including eye movement, eye blinks or heartbeats. We further downsampled the EEG recordings to 200 Hz, for consistency with the EOG-synthetically contaminated dataset. Finally, as will be done with all the datasets used in this work, we select the 19 standard channels from the international 10-20 electrode system and divide the dataset in two-second-long epochs.

We split the LEMON data into three subject-stratified groups: training (60\% of subjects), validation (20\% of subjects), and testing (20\% of subjects). We then populate the training, validation, and testing partitions by pooling the corresponding EEG epochs in each set.

\paragraph{Automatically denoised LEMON for Artifact Correction}
To train the implemented neural networks to automatically correct artifacts we need to build a dataset with noisy epochs as inputs and their corresponding cleaned epochs as targets. Regrettably, it is not possible to relate the raw and manually pre-processed signlas provided by Babayan et al. due to missing information about rejected data segments. Therefore, starting from the raw LEMON data, the input dataset, $\mathbf{\hat{X}}$, is obtained by downsampling the data to 200 Hz (for consistency with the rest of the used datasets) and applying a bandpass filter between 1 and 45 Hz. In parallel, we sample epochs from noisy recordings from the testing partition and merge them together with random epochs from the pre-processed LEMON testing partition, creating the "LEMON \textit{Clean/RawFiltered}" dataset mentioned in Section \ref{sec:ArtifactDetection}, to evaluate the artifact detection capabilities of the trained networks.
 
For the clean part, we repeat the downsampling and filtering of the raw data and later apply automatic rejection of Independent Components with ICLabel \cite{PionTonachini2019}. ICLabel is a deep-learning based tool that classifies independent components resulting from an ICA decomposition in several classes ("\textit{brain}", "\textit{muscle artifact}", "\textit{eye blink}", "\textit{heart beat}", "\textit{line noise}", "\textit{channel noise}", "\textit{other}"). We make use of the provided classes, along with the provided classification probability to reject independent components.

We study the following two automatic approaches:

- \textbf{Maintain only brain components}: only those independent components that have been labeled as "\textit{brain}" by the ICLabel tool with a probability over 80\% are kept, the rest are removed from the EEG signal, obtaining $\mathbf{X_{Br}}$, following the same approach as in \cite{Chuang2022}.

- \textbf{Reject artifactual components}: we also consider a more conservative approach where we only remove those independent components that have been labeled as artifact ("\textit{muscle artifact}", "\textit{eye blink}", "\textit{heart beat}", "\textit{line noise}", "\textit{channel noise}") with a probability over 90\%, obtaining $\mathbf{X_{Ar}}$. This is the default setting for the ICLabel plugin in the standard toolkit EEGLAB.

Again, we use the 19 standard channels in the international 10-20 electrode system and split the temporal dimension into two-second-long epochs.

\subsubsection{EOG-synthetically contaminated dataset}
The EOG-synthetically contaminated dataset consists of two different sets of data: one with clean EEG recordings, and one with EOG synthetically contaminated EEG recordings \cite{Klados2016}.
The clean EEG dataset, $\mathbf{X_{EOG}}$, contains 1 minute of eyes-closed (EC) EEG recordings from 27 healthy subjects (14 males) at a sampling frequency of 200Hz. A montage with the 19 standard electrodes was used, in the 10-20 electrode system. The recordings were already filtered by the authors, applying a band-pass filter in the range 0.5-40Hz and a Notch filter at 50Hz.
The synthetically contaminated set,  i.e. $\mathbf{\hat{X}_{EOG}}$, is obtained through the linear combination of EOG recordings with the clean EEG dataset. In this case, the EOG was recorded from the same subjects during an eyes-open (EO) task with four electrodes to obtain the vertical (VEOG) - through two electrodes - and the horizontal (HEOG) components - through the other two.
The EOG components are linearly combined with the clean EEG signals following:
\begin{equation}
    \mathbf{\hat{X}_{EOG}}^j=\mathbf{X_{EOG}}^j + a^j VEOG + b^j HEOG
\end{equation}
where $j$ is the channel index and $a^j$ and $b^j$ are linear coefficients that account for the strength contribution of the EOG to each EEG channel. These coefficients follow a model further explained in \cite{Klados2016}.

\subsection{LSTEEG Network Structure}

LSTM layers have been shown to perform well on physiological time signals \cite{Karim2018, Ravi2017}. The developed LSTM-based autoencoder takes multi-channel two-second-long EEG epochs as input, aiming to achieve unsupervised feature extraction of underlying EEG characteristics. The multi-channel epoch is first embedded into the Latent Space (LS) by the encoder section of the network. The encoder processes the epoch via two concatenated LSTM layers, with $N_o=50$ and $N_i=25$ features in the hidden state. The output of the LSTM layers is flattened and projected through a fully connected layer into the LS, of dimension $N_{LS}$. While the $N_o$ and $N_i$ have been fixed, $N_{LS}$ is left as free parameter since it is the parameter with the most influence on the final reconstruction capacity of the network (see Supp1.2 in the Supplementary Information).

The embedded information is then reconstructed through the decoder, effectively reversing the encoding process using a fully connected layer, two concatenated LSTM layers, now with $N_i=25$ and $N_o=50$ respectively, and a final fully connected layer that is charged to project from the $N_o \times N_T$ to the original $N_C \times N_T$.

\begin{figure}
    \centering
    \includegraphics[width=\linewidth]{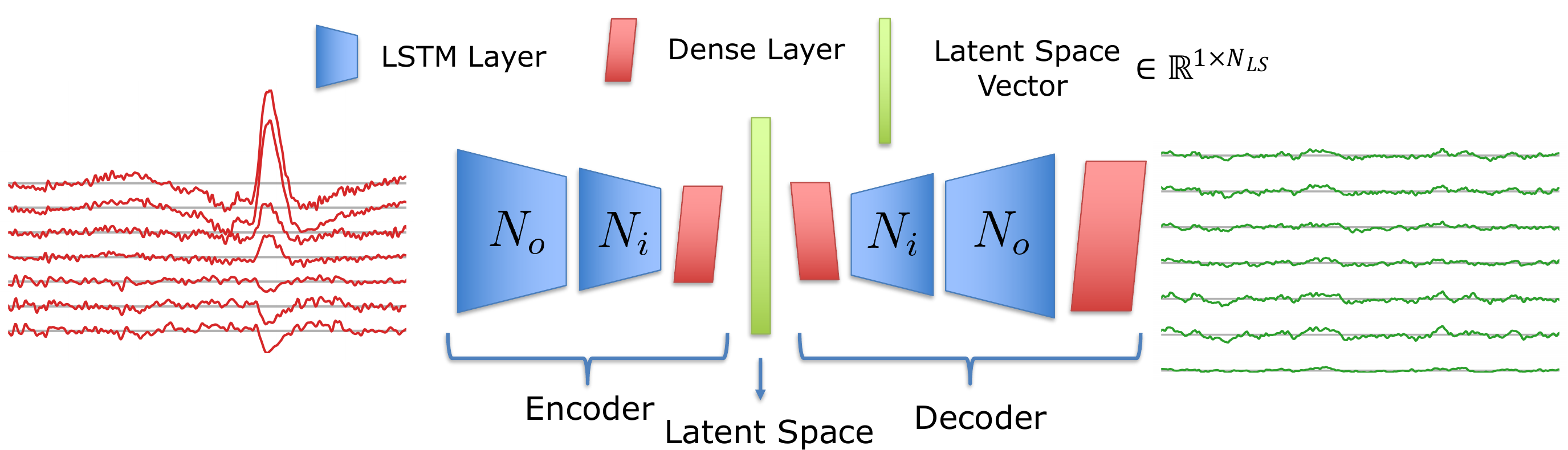}
    \caption{LSTEEG network architecture.}
    \label{fig:architecture}
\end{figure}

\subsection{Comparative Performance Evaluation with Convolutional Autoencoders}

Recent advancements in automatic EEG denoising have leveraged CNNs, achieving state-of-the-art performance. To benchmark our LSTEEG model against existing techniques, we implemented two leading convolutional autoencoders: the UNET architecture \cite{Chuang2022} and CLEEGN \cite{Lai2022}.

\subsubsection{UNET}
In \cite{Chuang2022} authors propose to adapt the UNET architecture to the denoising EEG problem. The UNET network is a popular Convolutional Neural Network (CNN) that is commonly used for biomedical image segmentation \cite{Ronneberger2015}. It has an AE structure (although not necessarily with information compression) in which the encoder network applies a series of convolution and downsampling operations while the decoder network applies a series of upsampling and convolution operations. For further details on the network implementation and performance on EEG denoising, please refer to \cite{Chuang2022}.

\subsubsection{CLEEGN}
The CLEEGN network is a lightweight convolutional autoencoder specifically developed for multi-channel inter-subject EEG reconstruction, that has strong decoding performances on BCI datasets \cite{Lai2022}. It encodes and decodes the input information through a series of 1D convolutional filters, both in the channel and time dimensions. However, no compression of information takes place between the encoder and the decoder as the Latent Space has $N_F$ times the dimension of the input, being $N_F$ the number of temporal filters. The reader may refer to the CLEEGN original publication \cite{Lai2022} for extended details.

\subsection{Latent Space Exploration}
During the training phase, LSTEEG's encoder learns to project the high-dimensional information in an EEG sample onto a lower-dimensional Latent Space in an unsupervised manner, allowing the model to autonomously learn relevant features and characteristics from the input data, i.e. from expected clean EEG. Therefore, probing the structure and characteristics of the learned LS can provide meaningful insight for data-driven knowledge discovery on the characterization of artefact-free EEG.

\subsubsection{Latent Space Dimension Activations}
We first aim to understand the activation patterns within the Latent Space to identify which EEG signal components most strongly influence each latent dimension. Specifically, the LSTEEG's encoder projects each EEG epoch ($x_e$)  onto the LS through the mapping $x_e \rightarrow f_E(x_e) \in \mathbb{R}^{N_{LS}}$. In this context we analyze the characteristics of $x_e$ that contribute to an increase or a decrease in the activation values at each dimension $j$ of the LS, $f_E^j(x_e)$.

Spectral features, which characterize the signal's power distribution across different frequency bands, have been widely applied to analyze resting state EEG \cite{Cohen2014, Koenig2009}. Furthermore, the multi-channel nature of our methodology can be leveraged to investigate how spatial distributions of spectral features influence LS activation patterns.

Therefore, to quantify the relationship between spectral features and LS activation, we calculate the relative band power for the five standard EEG bands ($b$) previously mentioned (Delta, Theta, Alpha, Beta, Gamma), across each channel ($c$) and epoch ($e$): $P_{b, c, e}$. We then construct an the activation map for each dimension in the LS through:

\begin{equation}
    S_{b,c}^j = \sum_{e=1}^{N_e} P_{b, c, e} f_E^j(x_e)
\end{equation}

where $S_{b,c}^j$ is the spectral activation for dimension $j$, band $b$, and channel $c$. A topographical map is subsequently generated for each dimension $j$ and band $b$. Each element's contribution to the activation is weighted by its corresponding LS encoding, $f_E^j (x_e)$, ensuring that the summation highlights patterns that produce the most significant activation values.

In this manuscript, we choose to visualize the Most Activated LS Dimensions ($MAD$s). These are the dimensions characterized by the highest cumulative activation across all epochs. The cumulative activation for one dimension of the latent space $j$ can be defined as $A_j = \sum_{e=1}^{N_e}\left|f_E^j (x_e)\right|$. Therefore, the set of K-Most Activated Dimensinons (MADs) is defined as:
\begin{equation}\label{eq:MADs}
    \text{MAD}_K = \{ j_1, j_2, \dots, j_K \, | \, A_{j_1} \geq A_{j_2} \geq \dots \geq A_{j_K} \}
\end{equation}
with $j_1, j_2, \dots, j_K \in \{1, 2, \dots, N_{LS}\}$. In this manner, we aim to focus the scope of the analysis on the features that most powerfully influence the neural representations within the model.

\subsubsection{Linear Interpolation in the Latent Space}
Linear interpolation within an autoencoder's LS is a powerful technique to assess the smoothness and generalization capabilities of the learned representations. By generating intermediate representations between two points in the LS and decoding them back to the input space, it is possible to determine whether the LS transitions smoothly and meaningfully between known data points. Smooth transitions are indicative of a model that is well-structured and has learned to generalize well across the data distribution \cite{Kingma_2013}, which is potentially beneficial for later downstream tasks and improved interpretability \cite{higgins2017betavae}. Finally, a regular and smooth LS allows to generate meaningful synthetic data, as points between actual samples retain realistic characteristics, which can be leveraged to expand datasets through data augmentation.   

We therefore attempt to verify the regularity of the LS by randomly selecting two epochs ($x_a$ and $x_b$), obtaining their LS projections ($f_E(x_a)$ and $f_E(x_b)$), and sampling intermediate vectors that lie in the segment connecting $f_E(x_a)$ and $f_E(x_b)$.

In mathematical terms, let $\lambda$ represent the interpolation parameter that varies linearly from 0 to 1. To obtain $M$ interpolated vectors we define $\lambda_m = \frac{m}{M}$, $ m = 1, 2, \ldots, M$. Then, the interpolated vectors, denoted as $z^{int}_m$, can be expressed as:

\begin{equation}
z^{int}_m = (1 - \lambda_m) \cdot f_E(x_a) + \lambda_m \cdot f_E(x_b)    
\end{equation}

Finally, these intermediate vectors represent are decoded $y^{int}_m = f_D(z^{int}_m)$ to visualize them in the EEG space, for comparison with the real EEG samples $x_a$ and $x_b$.

\section{Results}\label{sec:Results}

\subsection{Artifact Detection}
\begin{figure}[!ht]
    \centering
    \includegraphics[width=\linewidth]{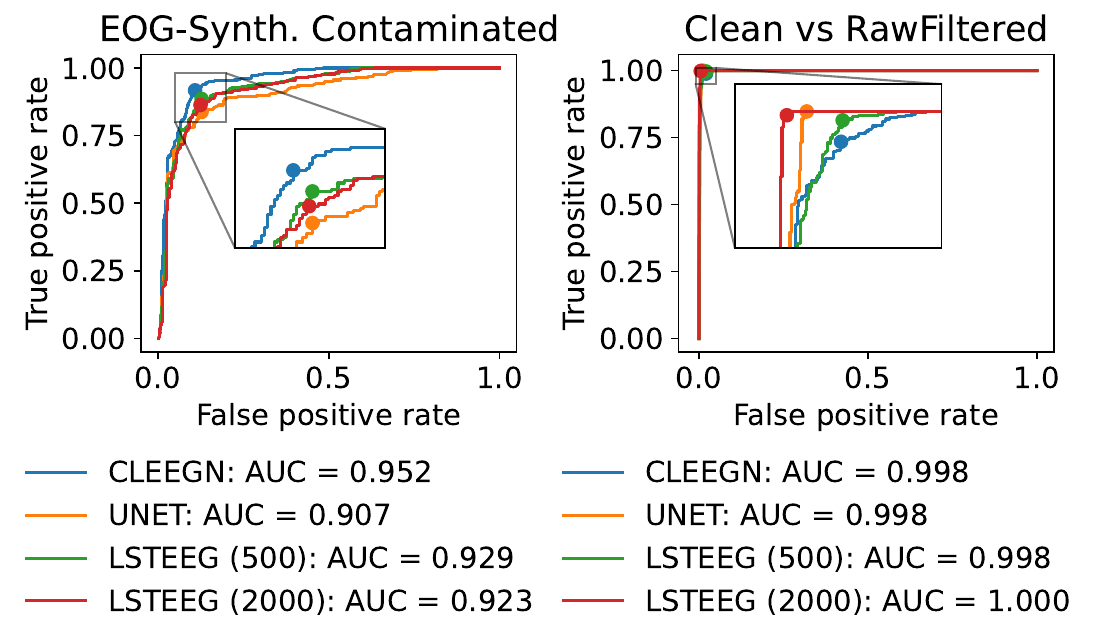}
    \caption{Artifact Detection performance of the CLEEGN, UNET and two different LSTEEG configurations ($N_{LS} = 500$ and $N_{LS}=2000$) on the two evaluation datasets, the EOG-synthetically contaminated dataset (left) and LEMON \textit{Clean/RawFiltered} dataset (right). We compare the different networks' classification performance by showing their Receiver Operating Characteristic (ROC) curves. Additionally, Area Under the ROC Curve (AUC, higher is better) values are provided in the figures legends.}
    \label{fig:ArtDet-Summary}
\end{figure}

Figure \ref{fig:ArtDet-Summary} shows the artifact detection ROCs when applying the four compared networks both on the EOG-synthetically contaminated (left panel) and the LEMON \textit{Clean/RawFiltered} (right panel) datasets. In both cases the networks have been trained on the pre-processed LEMON dataset. As it can be observed, the AUC values are generally large, always surpassing 0.9, and indicating that the anomaly detection approach classifying noisy against clean epochs works satisfactorily for the implemented autoencoders.

The LSTEEG ($N_{LS} = 2000$) and the UNET networks are the top performers on the \textit{Clean/RawFiltered} dataset, which is the evaluation dataset most similar to the training one. Meanwhile, on the EOG-Synthetically Contaminated dataset, recorded with a different EEG set up in a different laboratory, the CLEEGN and LSTEEG ($N_{LS} = 500$, but closely followed by $N_{LS} = 2000$,) networks perform best.

These results demonstrate that the novel anomaly-detection-based EEG artifact rejection method is a powerful algorithm for detecting artifacts in EEG epochs, effectively generalizing to datasets recorded under different conditions from the training data. The four tested networks consistently achieve remarkable performance across both datasets, highlighting the robustness of the method.

\subsection{Artifact Correction}

\begin{table}[ht]
\centering
\renewcommand{\arraystretch}{1.3}  
\resizebox{0.9\linewidth}{!}{
\begin{tabular}{|c|c|}
\hline

\rowcolor[HTML]{9B9B9B} 
$\mathbf{X_{Br}}$ & \textbf{Correction RMSE ($\mu V$)} \\ \hline
CLEEGN & 2.24 $\pm$ 5.58 \\ \hline
UNET & \textbf{1.60 $\pm$ 0.76} \\ \hline
LSTEEG ($N_{LS}=500$) & 1.78 $\pm$ 0.72 \\ \hline
LSTEEG ($N_{LS}=2000$) & 2.01 $\pm$ 0.75 \\ \hline

\rowcolor[HTML]{9B9B9B} 
$\mathbf{X_{Ar}}$ & \textbf{Correction RMSE ($\mu V$)} \\ \hline
CLEEGN & \textbf{1.56 $\pm$ 1.868} \\ \hline
UNET & 2.39 $\pm$ 9.67 \\ \hline
LSTEEG ($N_{LS}=500$) & 3.06 $\pm$ 10.86 \\ \hline
LSTEEG ($N_{LS}=2000$) & 2.41 $\pm$ 10.90 \\ \hline
\end{tabular}
}
\caption{MSE results reported as $\text{Mean} \pm \text{SD}$ over EEG epochs in the test partition for different networks trained with the two automatically pre-processed datasets $\mathbf{X_{Br}}$ and $\mathbf{X_{Ar}}$.}
\label{tab:ArtifactCorrMSEs}
\end{table}

We use the reconstruction error, i.e. the root mean squared error (RMSE) between the target epoch and the output of each network, as performance evaluation metric.
The reconstruction errors averaged over the EEG epochs in the testing set are shown in Table \ref{tab:ArtifactCorrMSEs}, for the four implemented networks when trained with the two different datasets discussed in Section \ref{sec:LemonDataset}. The UNET architecture is the outperforming model when training the networks with $\mathbf{X_{Br}}$, where we reject all independent components that ICLabel does not classify as \textit{brain} class with $p>0.9$. Our LSTEEG closely follows UNET in terms of performance, while CLEEGN shows both the largest and most variable error.

\begin{figure*}[ht]
    \centering
    \begin{subfigure}[b]{0.48\linewidth}
        \includegraphics[width=\linewidth]{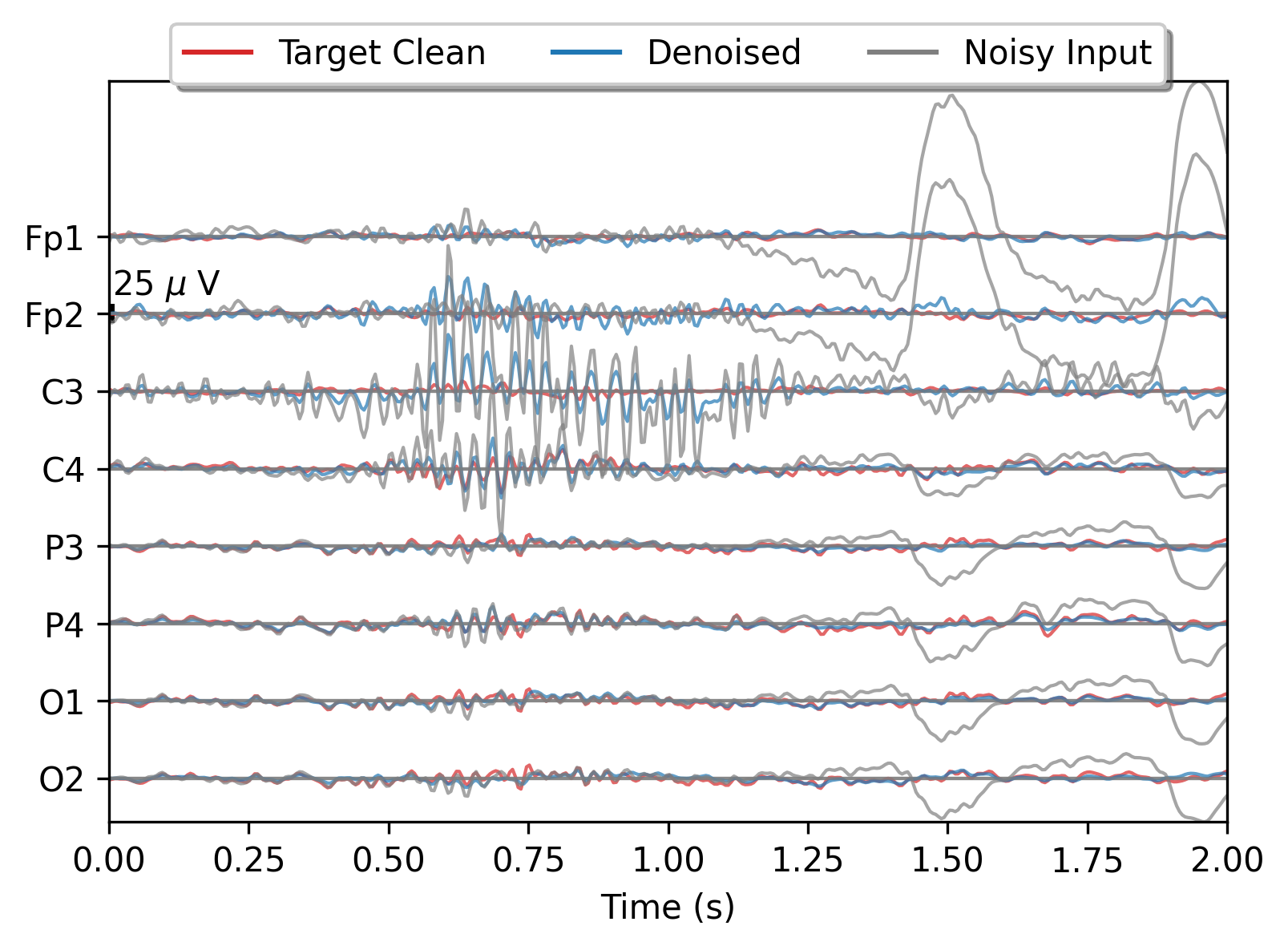}
        \caption{CLEEGN}
        \label{fig:Br_muscle_cleegn}
    \end{subfigure}
    \hfill 
    \begin{subfigure}[b]{0.48\linewidth}
        \includegraphics[width=\linewidth]{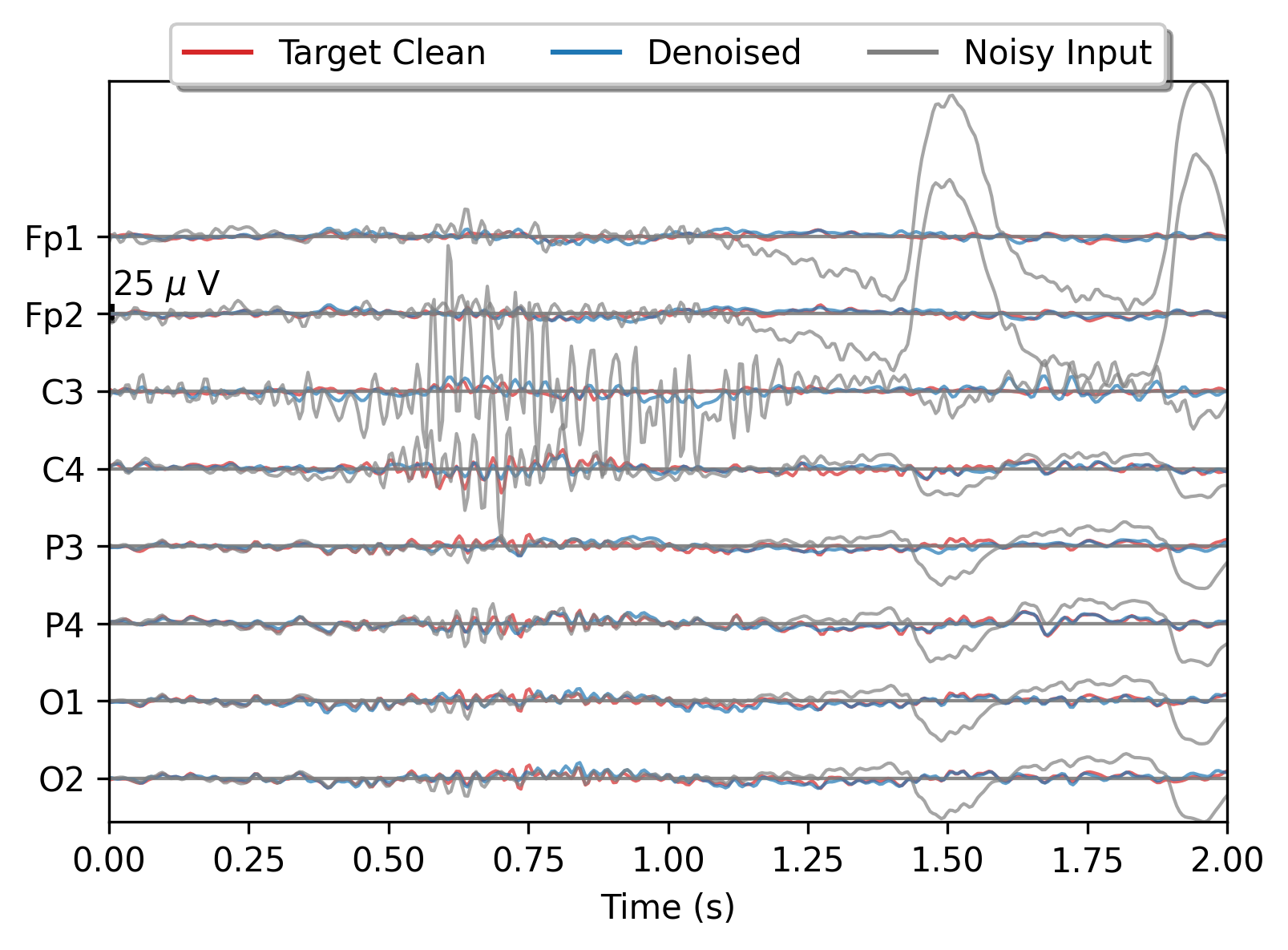}
        \caption{UNET}
        \label{fig:Br_muscle_unet}
    \end{subfigure}

    \begin{subfigure}[b]{0.48\linewidth}
        \includegraphics[width=\linewidth]{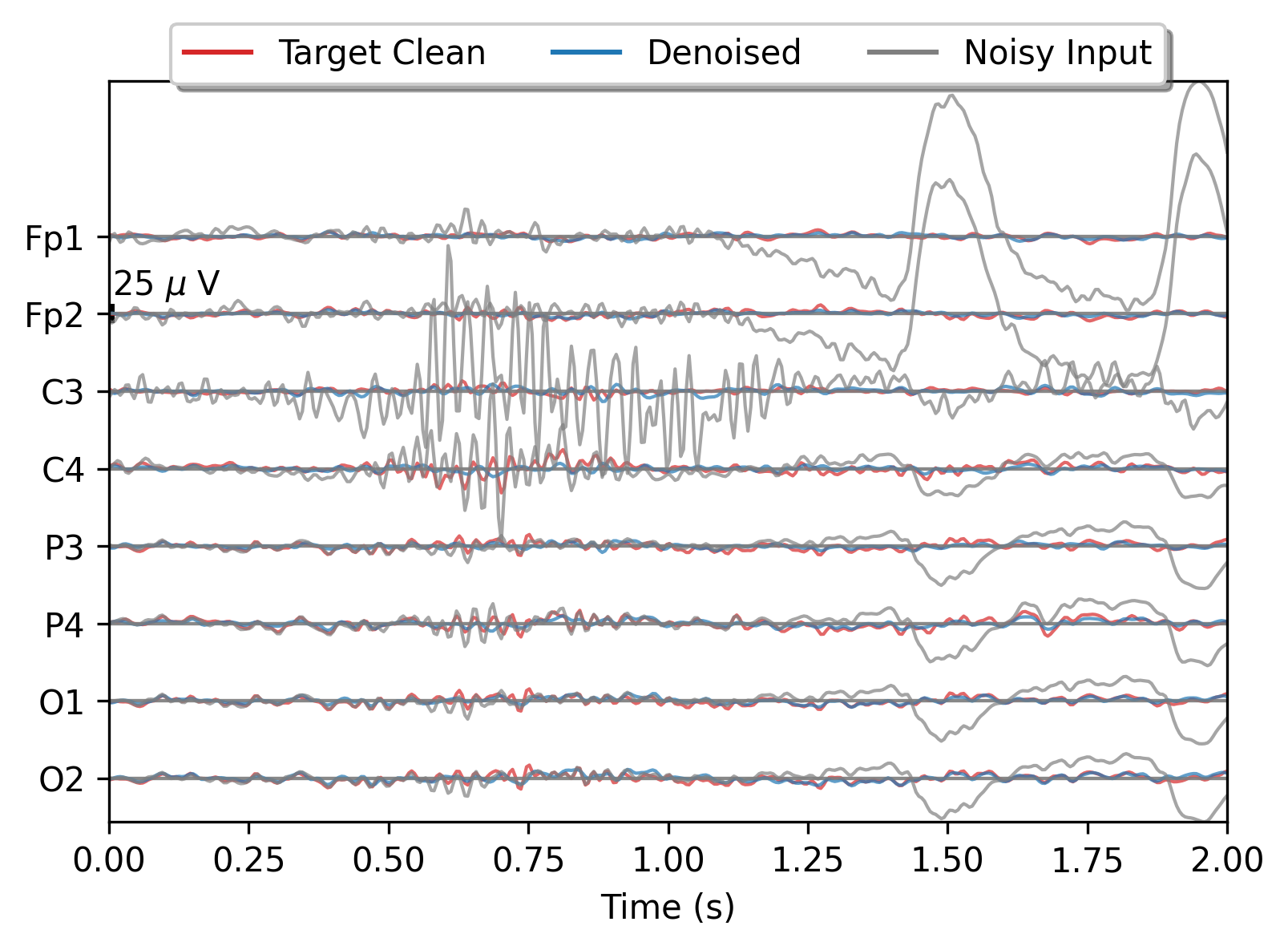}
        \caption{LSTEEG ($N_{LS}=500$)}
        \label{fig:Br_muscle_staraeeg500}
    \end{subfigure}
    \hfill 
    \begin{subfigure}[b]{0.48\linewidth}
        \includegraphics[width=\linewidth]{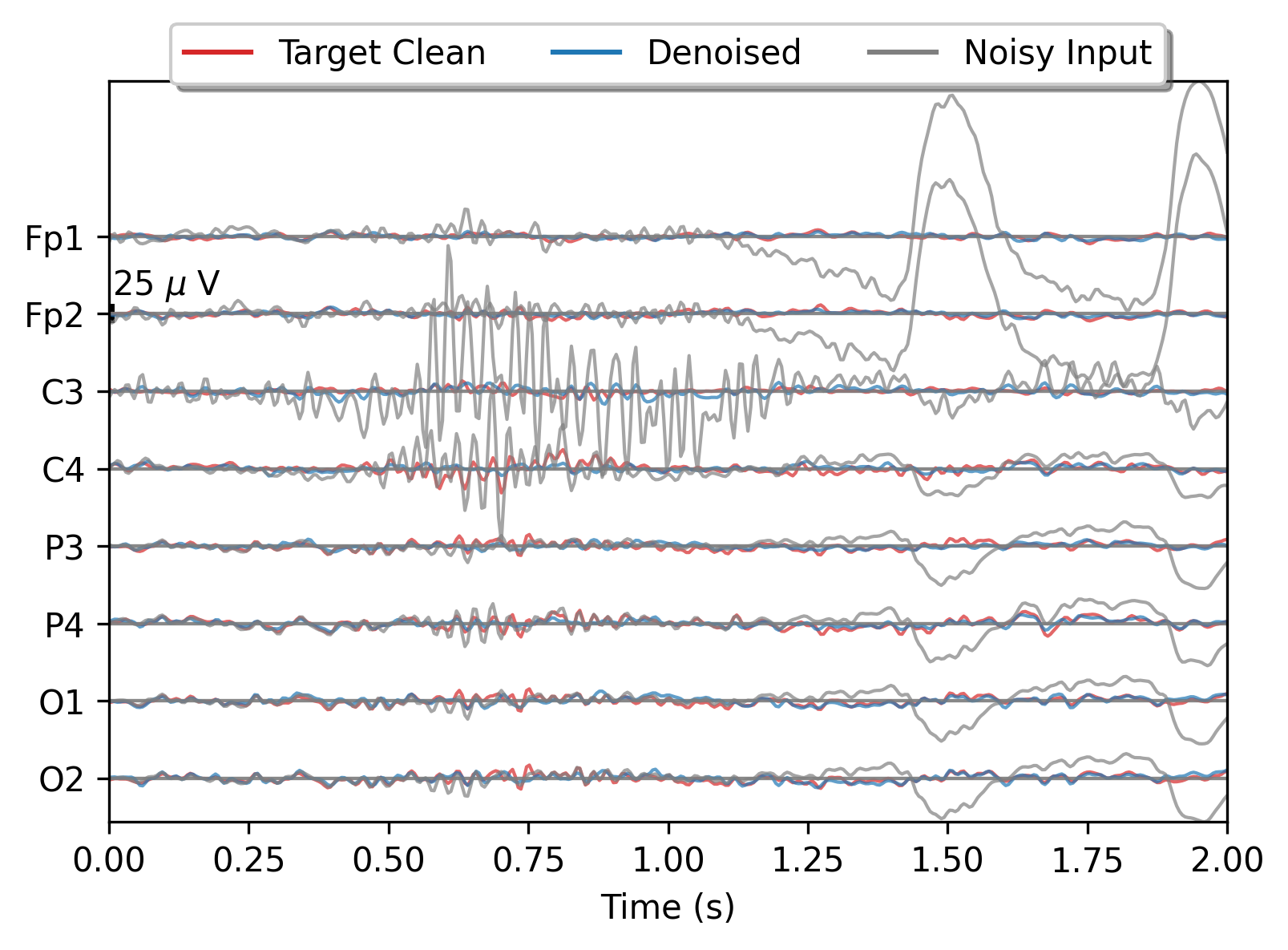}
        \caption{LSTEEG ($N_{LS}=2000$)}
        \label{fig:Br_muscle_staraeeg2000}
    \end{subfigure}

    \caption{Reconstruction Comparison for an epoch containing both muscular (high frequency burst between 0.5 and 0.75s) and ocular (high amplitude peaks at 1.50s) artifacts in the testing set. Shown networks trained with $\mathbf{X_{Br}}$. The Red line is the cleaned target EEG epoch; the Grey line is the input EEG epoch, containing the large amplitude artifact; the Blue line is the output of each network. While CLEEGN produces high amplitude outputs, unable to correct the artifact, both UNET and the two configurations of LSTEEG are able to remove the artifact from the EEG signal.}
    \label{fig:Br_muscle}
\end{figure*}

In Figure \ref{fig:Br_muscle}, a comparison of the correction capabilities for an EEG epoch containing a muscular and an ocular artifacts is shown. In the muscular artifact (high frequency burst of activity between 0.5 and 0.75s) one can see the difference in performance between the different networks: CLEEGN produces a denoised signal that is closer to the noisy input than the target signal. Meanwhile, LSTEEG and UNET produce outputs that follow closely the target signal, with the output of LSTEEG being slightly worse due to its smaller amplitude. 

When training the implemented models with the more conservative dataset $\mathbf{X_{Ar}}$, where only those independent components labeled as artifacts with high probability are rejected, results change significantly. Table \ref{tab:ArtifactCorrMSEs} shows the performance of the different models, where CLEEGN achieves the smallest reconstruction error, while UNET and LSTEEG ($N_{LS}=2000$) achieve almost identical performance. Additionally, CLEEGN achieves remarkably small variability in its RMSE values over the test set, while the rest of models show standard deviations of the order of 10 $\mu V$.

\begin{figure*}[ht]
    \centering
    \begin{subfigure}[b]{0.48\linewidth}
        \includegraphics[width=\linewidth]{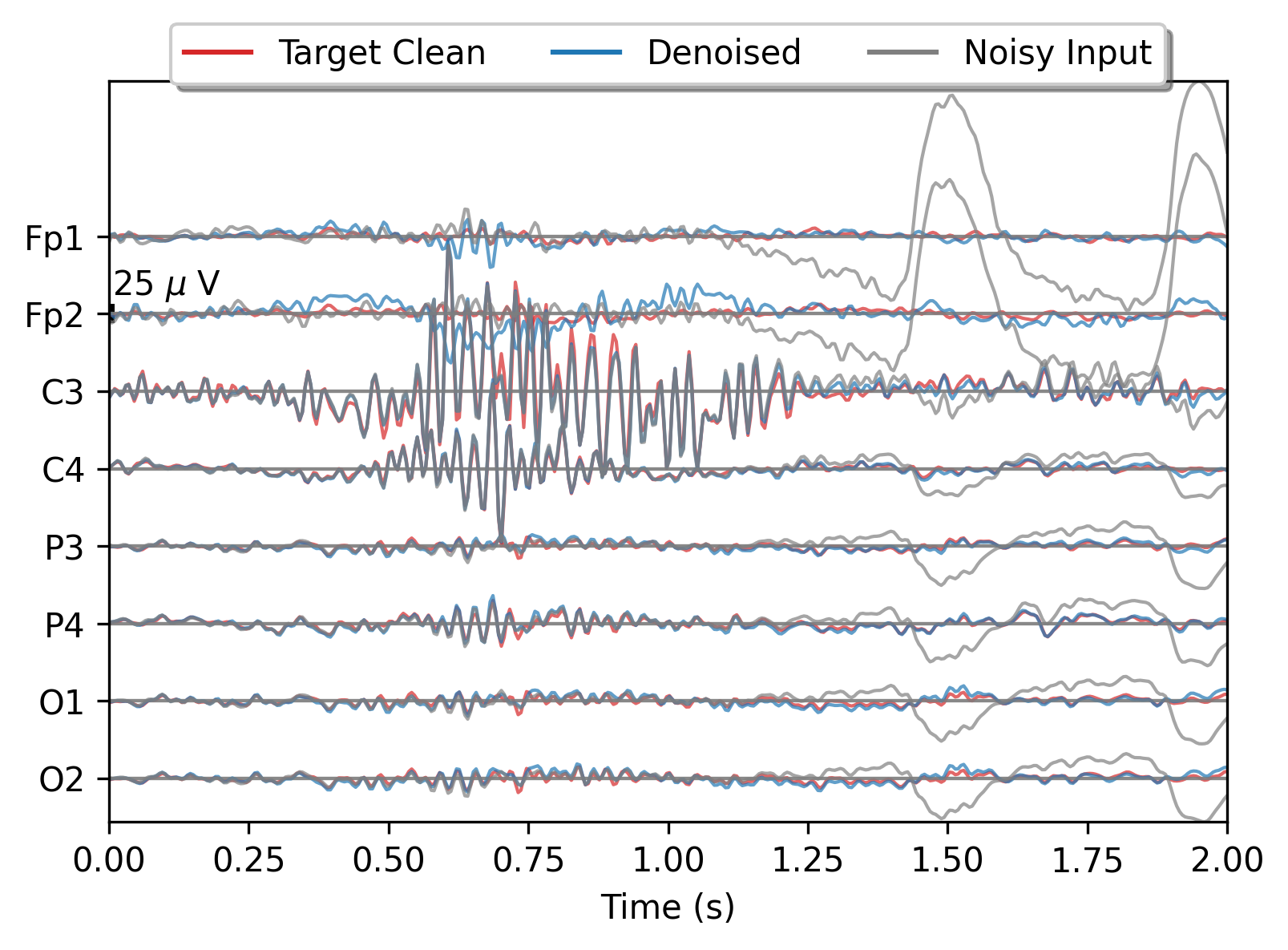}
        \caption{CLEEGN}
        \label{fig:AR_muscle_cleegn}
    \end{subfigure}
    \hfill 
    \begin{subfigure}[b]{0.48\linewidth}
        \includegraphics[width=\linewidth]{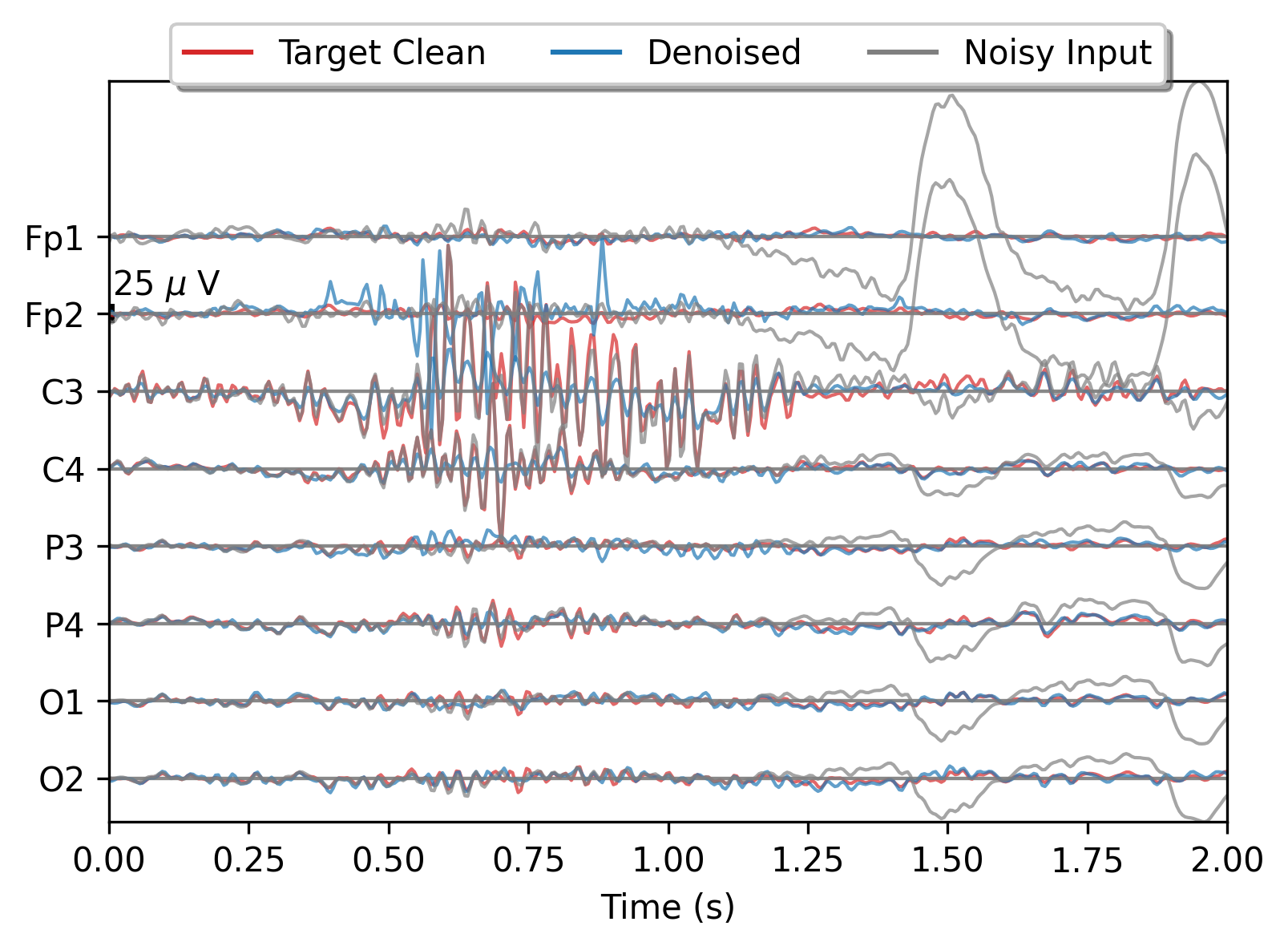}
        \caption{UNET}
        \label{fig:AR_muscle_unet}
    \end{subfigure}

    \begin{subfigure}[b]{0.48\linewidth}
        \includegraphics[width=\linewidth]{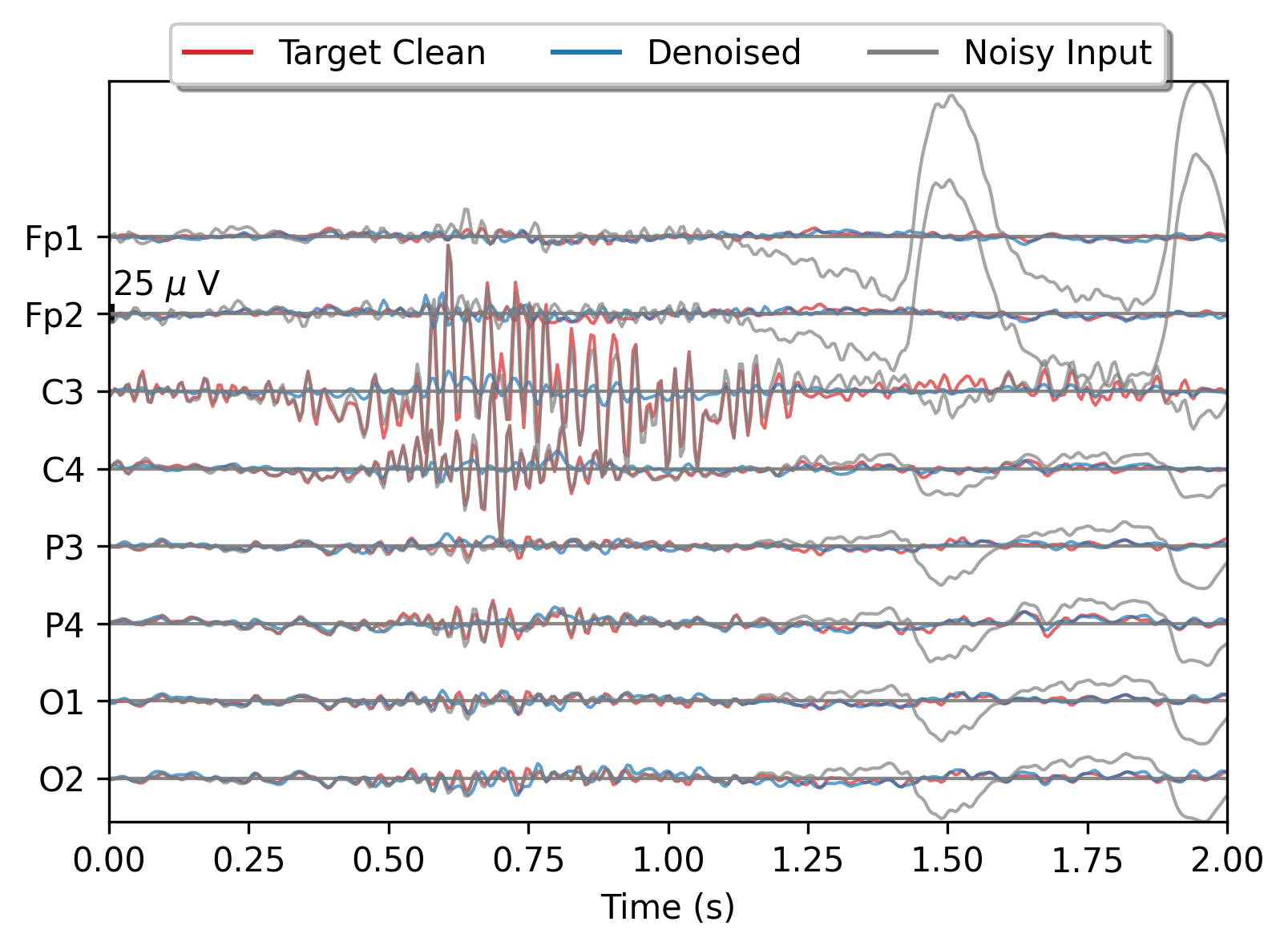}
        \caption{LSTEEG ($N_{LS}=500$)}
        \label{fig:AR_muscle_staraeeg500}
    \end{subfigure}
    \hfill 
    \begin{subfigure}[b]{0.48\linewidth}
        \includegraphics[width=\linewidth]{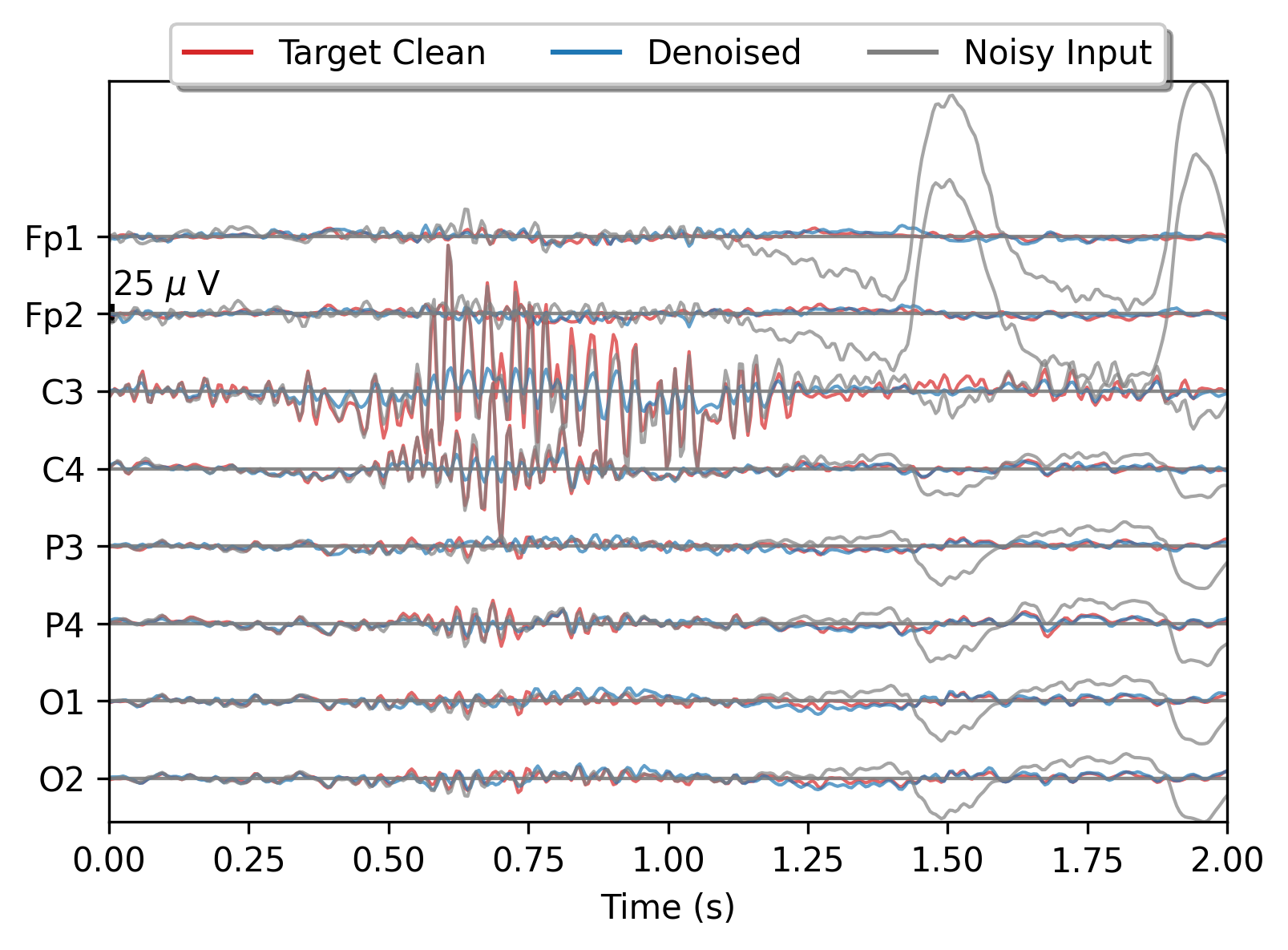}
        \caption{LSTEEG ($N_{LS}=2000$)}
        \label{fig:AR_muscle_staraeeg2000}
    \end{subfigure}

    \caption{Correction comparison for the same EEG epoch with muscular and ocular artifacts as in \ref{fig:Br_muscle}. Shown networks trained with $\mathbf{X_{Ar}}$. The Red line is the cleaned target EEG epoch; the Grey line is the input EEG epoch, containing the large amplitude artifact; the Blue line is the output of each network. While CLEEGN produces high amplitude outputs, unable to correct the artifact, both UNET and the two configurations of LSTEEG are able to remove the artifact from the EEG signal.}
    \label{fig:AR_muscle}
\end{figure*}

Such a difference in performance between the two datasets requires further investigation. Therefore, we visualize the EEG epoch previously discussed, now on the $\mathbf{X_{Ar}}$ dataset, and compare the reconstruction of the different models in Figure \ref{fig:AR_muscle}. One can easily interpret the quantitative results by inspecting Figure \ref{fig:AR_muscle}: in the $\mathbf{X_{Ar}}$ dataset, the \textit{"Target Clean"} signals may be noisier, due to taking a more conservative approach with ICLabel and rejecting less, possibly artifactual, Independent Components. Although the number of epochs with a contaminated target signal may be small, they are enough to degrade the performance of the CLEEGN model, which learns to closely match the target epoch independently of its form or shape. However, the remaining implemented networks seem to be able to have learned to generalize better from the large corpus of clean epochs in $\mathbf{X_{Ar}}$, and are able to correct the muscular artifact, thus producing larger RMSE values between the network output and the target epoch.

Finally, by inspecting the dimensions of each layer's output in the implemented networks, we argue that the previously discussed behavior is a clear result of how feature learning takes place in each network. Hence looking at the dimensions at the output of each of the networks in Table \ref{tab:outputs} for a two-second-long input epoch of size $(N_C, N_T) = (19, 500)$, the output dimensions of CLEEGN and UNET are always larger than the input dimension, implying a lack of information compression, which hinders information-rich low-dimensional representations in the Latent Space, and therefore points to the absence of feature embedding and representation learning.

\begin{table}[]
\resizebox{\linewidth}{!}{%
\renewcommand{\arraystretch}{1.5}  
\begin{tabular}{|c|c|c|c|c|}
\hline
\rowcolor[HTML]{C0C0C0} 
\textbf{Architecture} &
  \begin{tabular}{@{}c@{}}\textbf{Encoder Output}\\\textbf{Shape}\end{tabular} &
  \textbf{$\mathbf{N_{LS}}$} &
  \begin{tabular}{@{}c@{}}\textbf{CR}\\($\mathbf{N_{inp} / N_{LS}}$) \end{tabular} &
  \cellcolor[HTML]{C0C0C0}\textbf{\# Weights} \\ \hline
\textbf{CLEEGN}                   & (20, 19, 400) & 152000  & 0.05 & 16,541     \\ \hline
\textbf{UNET}                     & (512, 50)     & 25600   & $\approx$ 0.297  & 2,664,211  \\ \hline
\begin{tabular}{@{}c@{}}\textbf{LSTEEG} \\\textbf{($N_{LS}=500$)}\end{tabular}  & (500, )       & 500     & 15.2 & 10,053,969 \\ \hline
\begin{tabular}{@{}c@{}}\textbf{LSTEEG} \\\textbf{($N_{LS}=2000$)}\end{tabular} & (2000, )      & 2000    & 3.8 & 40,055,469 \\ \hline
\end{tabular}%
}
\caption{Table summarizing relevant dimensionality aspects of the implemented networks. The size of the Latent Space, $N_{LS}$, is the size of the flattened Encoder output. The compression ratio (CR) measures the relationship between the number of dimensions in the input space $N_{inp}$ (equal for all networks with $N_{inp} = 7600$), and the size of the Latent Space $N_{LS}$. The last column indicates the total number of trainable parameters in the networks.}
\label{tab:outputs}
\end{table}

\subsection{Interpretability of the Learned Latent Space}
The latent space is therefore worth studying due to the presence of representation learning achieved in LSTEEG. Moreover such a study can confer the architecture with explainability features that have not been exploited in other networks heretofore. The results shown in Figure \ref{fig:LS_Spatial} and Figure \ref{fig:LS_interpolation} correspond to LSTEEG ($N_{LS} = 500$), trained for anomaly detection, only with pre-processed and manually clean data, in order to visualize clean-eeg-like characteristics.

\subsubsection{Spatial Activations of the Latent Space Dimensions}
We highlight the interpretability capabilities of the proposed LSTEEG autoencoder's LS by showcasing the spectral activation maps for the four Most Activated Latent Space Dimensions (MADs) described in equation \ref{eq:MADs}.   

\begin{figure}[ht]
    \centering
    \begin{subfigure}[b]{0.48\linewidth}
        \includegraphics[width=\linewidth]{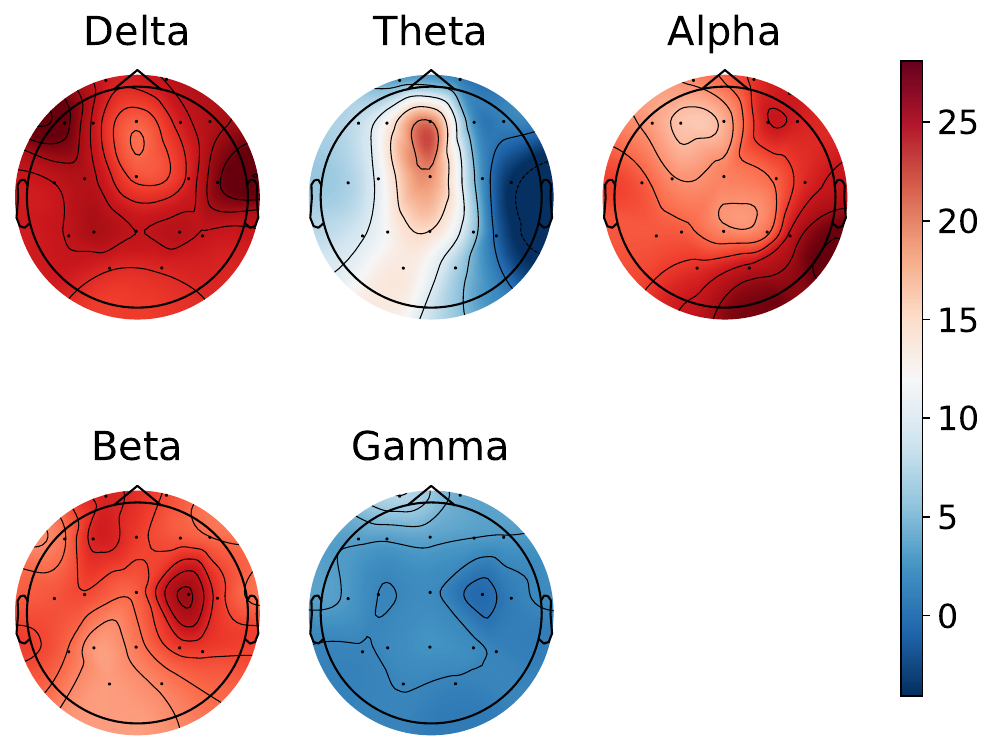}
        \caption{First MAD}
    \end{subfigure}
    \hfill 
    \begin{subfigure}[b]{0.48\linewidth}
        \includegraphics[width=\linewidth]{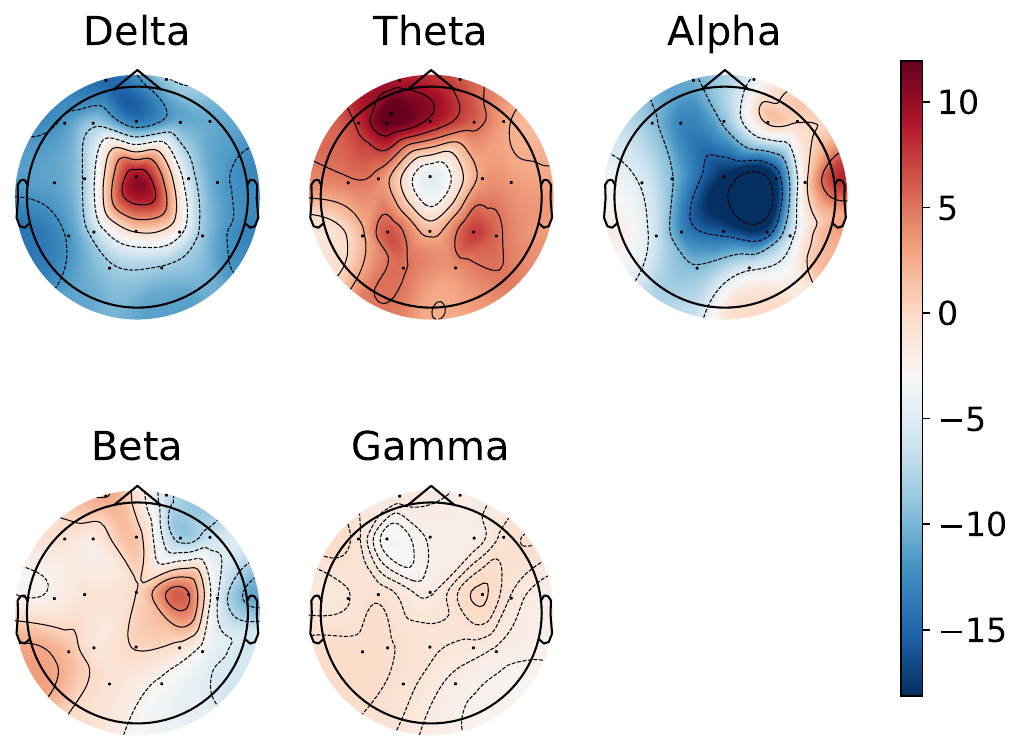}
        \caption{Second MAD}
    \end{subfigure}

    \begin{subfigure}[b]{0.48\linewidth}
        \includegraphics[width=\linewidth]
        {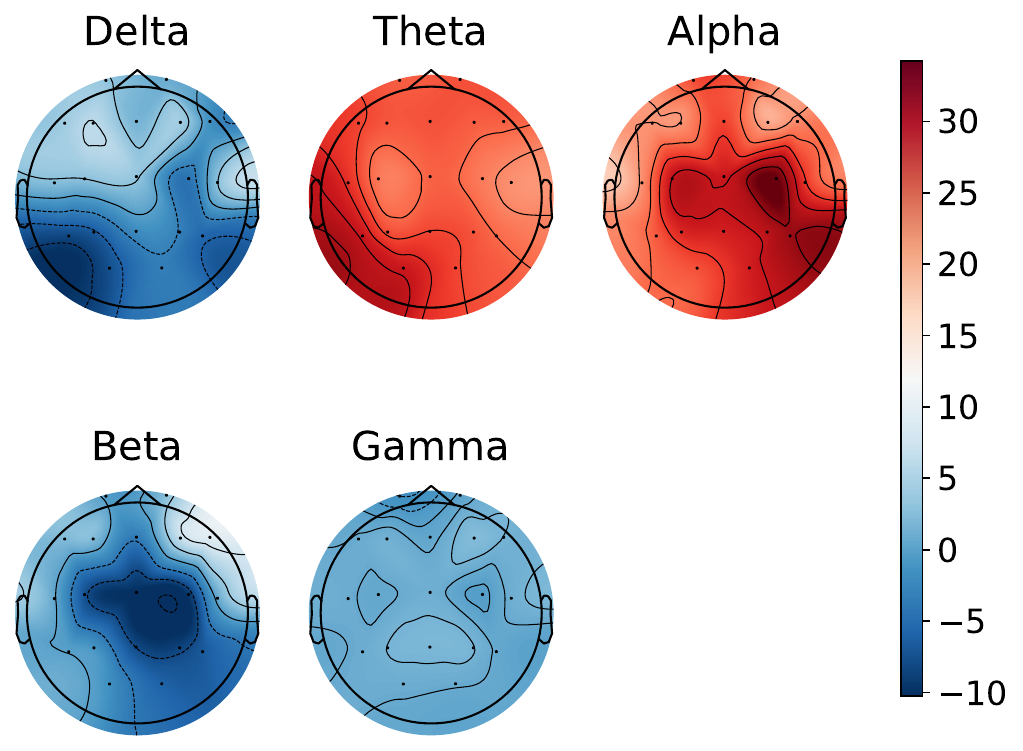}
        \caption{Third MAD}
    \end{subfigure}
    \hfill 
    \begin{subfigure}[b]{0.48\linewidth}
        \includegraphics[width=\linewidth]{Figures/LSExploration/mad3_lsdim.pdf}
        \caption{Fourth MAD}
    \end{subfigure}
    \caption{Topographical maps indicating activation patterns of Most Activated Latent Space Dimensions (MADs) by spectral features. }
    \label{fig:LS_Spatial}
\end{figure}

The activation maps of the MADs shown in  Figure \ref{fig:LS_Spatial} highlight some of thefeatures of the autoencoder that have been learned without supervision. One can readily notice that both the Delta and Alpha bands, closely followed by the Theta band, consistently exhibit large influences on the shown MADs, with clearly highlighted patterns. Meanwhile, the faster Gamma band tends to show smoother activation maps of much smaller amplitude.

This highlights the effectivity of this method at detecting different patterns of activity in the LS dimensions. The model does not prioritize encoding fast frequency information, instead focusing on encoding lower-frequency characteristics in the LS, an effect further analyzed in the discussion section.

While gamma band information is lost, we can see that each dimension encodes clearly defined spectro-spatial features, such as general delta, alpha and beta, power (0th), mid-central Delta (1st), central delta and alpha (2nd), etc. It is also worth noting that spatial activations are not simplified: dimensions are not specialized for particular patterns or particular band powers. Instead, each dimension  accumulates information from different band powers.

This methodology underscores the interpretative potential of EEG-based autoencoders. In this study, our developed algorithm offers insights into understanding the fundamental features of clean EEG signals \cite{Daly2012}, and it holds promise for application on a wider range of datasets to elucidate key characteristics associated with brain-related diseases or cognitive processes.

\subsubsection{Linear Interpolation in the Latent Space}
\begin{figure}[!ht]
    \centering
    \includegraphics[width=\linewidth]{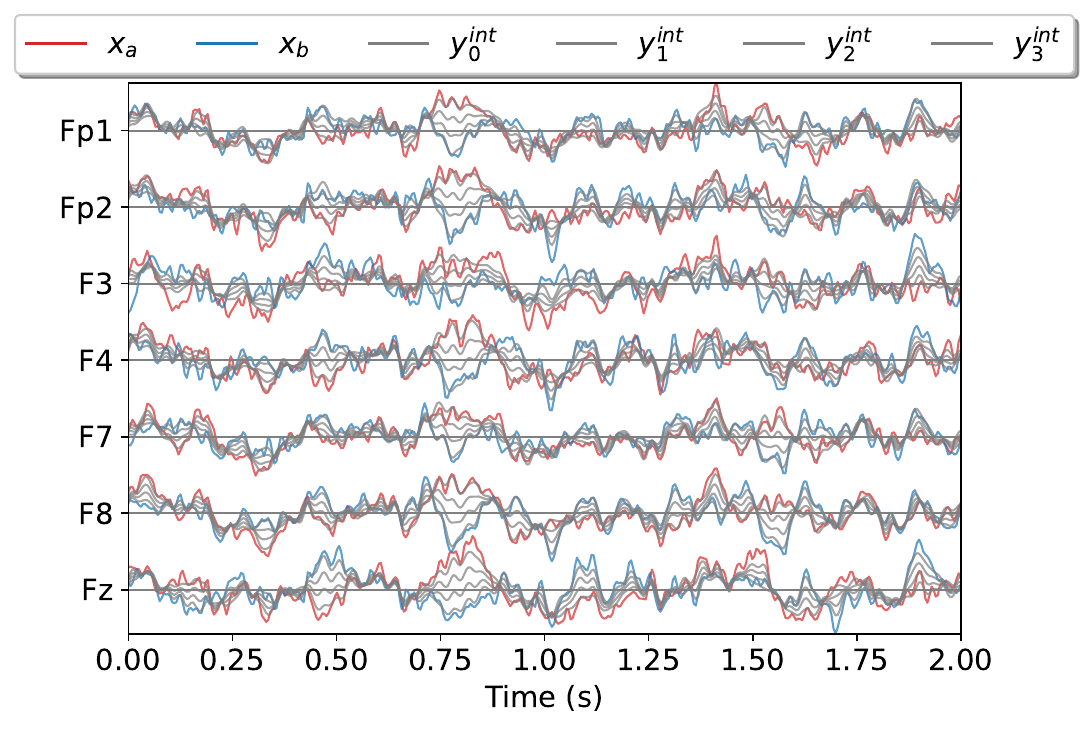}
    \caption{Interpolation of samples in the Latent Space. Regularity of the LS, allowing for generation of samples.}
    \label{fig:LS_interpolation}
\end{figure}

The regularity of the learned LS is explored by decoding the sampled points from linearly interpolation between two embedded EEG epochs onto the LS. Figure \ref{fig:LS_interpolation} shows the decoded samples in EEG space.

The interpolated points exhibit a smooth transition from sample A to sample B, the decoded interpolation points being confined between the two original samples' amplitudes. These results provide empirical evidence that the learned LSTEEG's LS is smooth, robust and generalizable. These features reveal LSTEEG's potential to be exploited as a data augmentation tool. Interpolating samples between two noisy EEG signals can be useful to generate synthetically augmented and labelled datasets to improve deep learning techniques for automatic artifact detection. Furthermore, the structured LS can be leveraged to generate additional EEG samples from individuals, by interpolating the LS between their real samples, to enlarge datasets aiming to enhance classification algorithms for e.g. brain-related disease diagnosis, patient stratification, emotional recognition, or BCI applications. 

\section{Discussion}\label{sec:Discussion}
In this study, we present LSTEEG, our novel LSTM autoencoder, and its performance on two EEG artifact-related problems: artifact detection and artifact correction, showcasing on par performance in artifact correction and outperforming in artifact detection existing state-of-the-art methodologies. The principal strength of LSTEEG is its distinctly structured Latent Space, which opens up a window of exploration into the underlying processeses of artifact detection and correction.

Additionally, we have successfully demonstrated the efficacy of the proposed anomaly-detection-based approach for artifact detection. This approach can be leveraged to improve ICA convergence by removing time segments with sporadic artifactual activity, which usually hinders ICA decomposition, aimed to identify stereotypical artifactual sources. Furthermore, the proposed methodology for noisy epoch removal can complement any artifact correction approach, using it as a last step to ensure cleanliness of EEG data.

In the context of artifact correction, we have compared LSTEEG with two state of the art Convolutional AEs: CLEEGN and UNET. Our analyses demonstrate that LSTEEG and UNET consistently show similar and powerful denoising performances. Both networks are able to remove different types of noise in EEG epochs such as ocular and muscle artifacts, as well as high amplitude deflections, occasionally caused by movement in the EEG electrodes. Additionally, their correct performance is robust against datasets that contain leaked artifactual activity in the target epochs. In such cases, the CLEEGN network learns to correct the most frequent artifacts, such as eye blinks, but does not successfully remove sporadic artifacts. 

Due to its network architecture and absence of information funnel, CLEEGN does not aim to learn a meaningful representation of EEG signals during training, but instead learns to very accurately transform the input EEG segment into the desired target. It is, therefore, less robust to noise in the training data. Instead, the encoder-decoder structure of LSTEEG and UNET encourages the networks to learn the underlying features characterizing clean EEG recordings, allowing for a more robust noise correction, even when the target epochs contain incorrectly removed artifactual activity from incorrectly classified ICs.

Nonetheless, CLEEGN is an extremely light model configuration, with competitive performance, while our LSTEEG models are significantly larger. We believe that, in addition to the increased robustness against noisy datasets, the enhanced interpretability and smaller dimensionality of LSTEEG's LS justifies the larger size of our novel autoencoder. LSTEEG's LS improves interpretability and extends the applicability of the LSTEEG model to other areas of EEG research. Possible future applications exploiting the low-dimensional embedding include EEG data generation and unsupervised EEG feature learning. 
The proposed model can be straightforwardly extended into a beta-variational autoencoder, potentially enhancing the smoothness of the latent space to facilitate data generation processes and interpretability by disentangling the dataset's generative factors \cite{Kingma_2013, higgins2017betavae, VAE_tutorial_kingma}. Complementarily, our studies show that transitions between points in the LS are relatively smooth, even without explicit regularization of the space. Finally, the low-dimensional LS embeddings can serve as an unsupervised EEG feature extraction head to enhance later downstream tasks such as brain-related disease diagnosis \cite{Ruffini2018, Pirrone2022}, EEG-based emotional recognition \cite{Akhand2023}, or seizure detection \cite{Liang2010, AbouAbbas2022}. 

We have shown that our multi-channel approaches to artifact detection and correction provide satisfactory performances. The selected configuration is widely applicable to EEG recordings adhering to the commonly used 19-channel setup from the 10-20 system. 

Furthermore, our work introduces a novel effective EEG artifact detection methodology based on anomaly detection. However, defining a reconstruction error threshold, over which an epoch is considered as artifactual, is not trivial. For instance, a dataset recorded with a different EEG device to the one used to collect training data may have different amplitude values, increasing baseline reconstruction error and degrading artifact detection performance. Therefore, normalizing the data before processing them through the deep neural networks is of utmost importance. The eventual selection of the detection threshold will depend on the position of the network in the pre-processing pipeline when combining it with other methodologies. Hence, it is recommended to allow less or more restrictive artefact rejection depending if it is respectively at the beginning or at the end of the pipeline.

In the context of artifact correction, it has become evident that our automatic methods for compiling the training dataset generate possibly noisy target data. Despite the robustness that LSTEEG and UNET have demonstrated against noisier training targets, as exemplified in Figure \ref{fig:AR_muscle}, obtaining a large and varied manually cleaned EEG dataset would largely improve the performance of the denoising techniques. Including different EEG tasks (resting state, motor, cognitive, sensory processing tasks, evoked potentials, etc.) in the training dataset would also widen the applicability range of the described networks to other experimental paradigms.

Nonetheless, deep learning methods that readily  correct EEG epochs offer a strong alternative to ICA-based  approaches \cite{Chuang2022}. Unlike ICA, which requires long recording segments to achieve adequate decomposition and can be affected by spurious artifacts, the proposed techniques can denoise individual epochs independently. This method significantly enhances the efficiency of correcting large recordings by allowing parallel processing of epochs through GPU-accelerated procedures using the inference phase of autoencoders. This parallelization not only speeds up the process but also improves the scalability of EEG signal correction.

Finally, despite their top performance, we have observed that LSTEEG and UNET exhibit a consistent attenuation in the PSD with increasing frequencies (see section Supp1.5 in the Supplementary Information). This seems to be a problem shared by different deep learning methods proposed for automatic EEG artifact correction \cite{Chuang2022, Brophy2022}. There are two inter-connected effects that encourage such an attenuation: the PSD's decaying spectrum in EEG signals and the Spectral Bias of neural networks. On the one hand, as has been widely reported, EEG signals exhibit a decaying power spectrum, usually following a power law \cite{Ferree2003, Miller2009}. The training goal of the implemented autoencoders is to minimize the MSE between the input and the network's output, which biases the learning towards the slowest frequencies. This is due to these bands carrying the largest relative power percentage in the signal, and therefore they predominantly affect the value of the MSE. On the other hand, Spectral Bias is a phenomenon that occurs in neural network training where lower frequency functions are learned before higher frequency ones \cite{Rahaman2019}. During the initial phases of  training, AEs tend to encode lower frequency information within the LS dimensions. As training progresses, and the model begins to learn higher frequency functions \cite{Rahaman2019}, fewer LS dimensions remain available to represent this information. This, coupled with the MSE loss function's emphasis on accurately reconstructing the dominant, lower frequency components, leads to a degradation in the representation capability for higher frequencies.

Overall, our LSTM autoencoder, characterized by its innovative design and the informative nature of its latent space, not only competes with contemporary methodologies but also provides a versatile foundation for varied applications in EEG data analysis. This study's findings open new avenues for research and development in understanding and processing EEG data, marking a significant contribution to the field.

\section{Conclusion}\label{sec:Conclusion}
In this study, we have introduced LSTEEG, a novel LSTM autoencoder designed for EEG artifact detection and correction. Thus we have  introduced a new approach that allows to leverage unsupervised learning with autoencoders for artifact detection in EEG epochs, where our implementation shows more robust performance than current state of the art convolutional autoencoders on two different datasets.

Moreover LSTEEG demonstrates comparable correction performance to current state-of-the-art methods and even outperforms them in some settings. Furthermore, it offers additional advantages through its structured Latent Space (LS) resulting from the representation learning process. This structured LS can not only be applied to enhance the interpretability of the EEG data but also opens up new research opportunities in synthetic EEG data generation and unsupervised feature learning.

\section*{Acknowledgment}
We would like to acknowledge Tarandeep Singh and Arnau Saumell for initial code implementations and development. We thank Giulio Ruffini for helpful and insightful discussions.

\bibliographystyle{ieeetr}
\bibliography{snbibliography}

\end{document}


\title[Article Title]{EEG Artifact Detection and Correction with Deep Autoencoders}

\section{Supplementary Information}\label{sec:SupInfo}
\subsection{Network Training Details}\label{sec:Supptraining}
All networks have followed the same training pipeline and same hyperparameters. Network convergence has been ensured by visual inspection of the learning curves, visualizing the evolution of both the training and validation losses.
Both the UNET and CLEEGN architectures have been trained as provided by the source code reported in the original publications \cite{Chuang2022, Lai2022}, respectively. We applied dropout after each layer in our LSTEEG architecture, with a probability of $p=0.1$.

All networks have been trained from scratch, with default weight initialization, for N=1000. The training loss has been used for weight updating, through backpropagation, using Adam optimizer \cite{Kingma_adam} with an initial learning rate of $\eta = 5\times 10^{-4}$ and a Cosine Annealing learning rate scheduler with $T_{max}=10$. Early stopping based on the validation loss was used. The selected batch size was 16, and the loss function, as described in the main text, the MSE between the target (either noisy input or denoised target) and the network’s output.
Hyperparameters have been adapted to the LEMON dataset.

\subsection{Exploration of LSTEEG parameters}\label{sec:Supp_params}

Initial, preliminary, hyperparameter explorations showed that, while small values of $N_o$ and $N_i$ may deteriorate the performance of the autoencoder, the determining factor in improving the reconstructing capabilities of the network is the size of the LS, $N_{LS}$. Therefore, we chose to fix $N_o=50$ and $N_i=25$. These two values neared the elbow in reconstruction error in the two studies presented in Figures A and B, while keeping the total number of parameters in the network to a reasonable value.

Finally, in Figure C, it is shown that, for the chosen $N_o$ and $N_i$ values, the dimension of the Latent Space has a strong effect in the reconstruction error of the network. The smaller the LS, the more compression, the higher the loss of information, which of course degrades the reconstruction and increases the error.

\begin{figure}[ht]
    \centering
    \begin{minipage}[b]{0.9\textwidth}
        \centering
        \includegraphics[width=\textwidth]{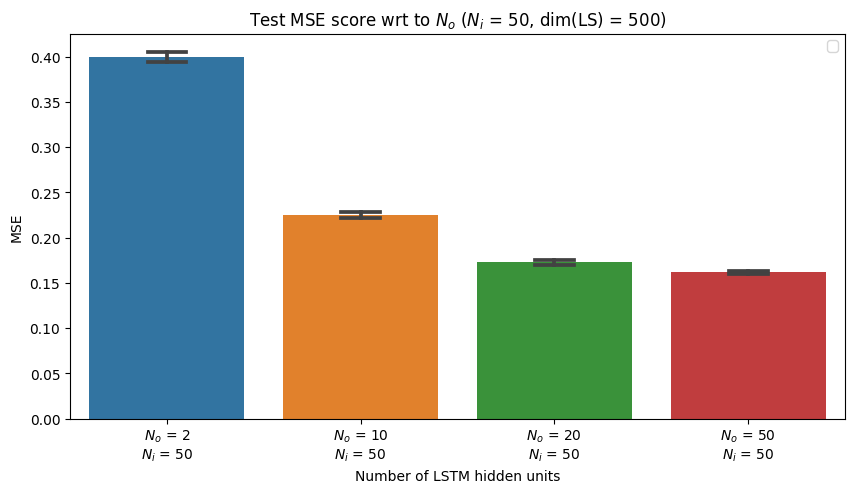}
        \caption*{(a)}
    \end{minipage}
    \hfill
    \begin{minipage}[b]{0.9\textwidth}
        \centering
        \includegraphics[width=\textwidth]{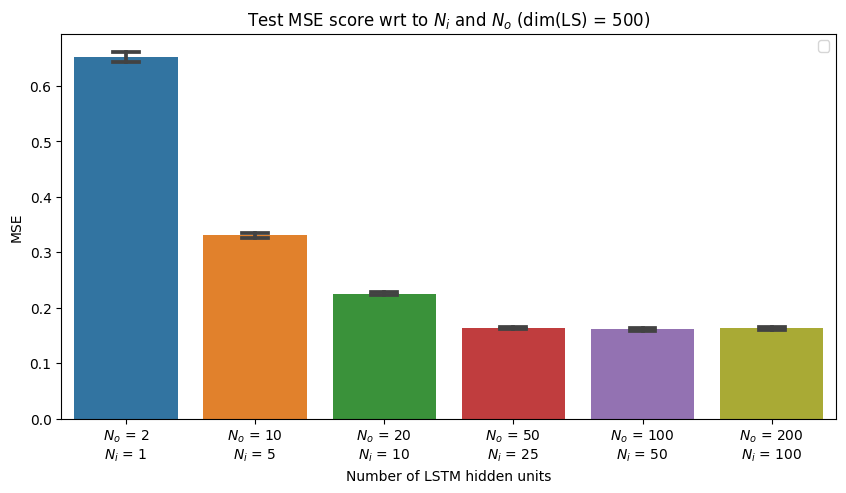}
        \caption*{(b)}
    \end{minipage}

    \begin{minipage}[b]{0.8\textwidth}
        \centering
        \includegraphics[width=\textwidth]{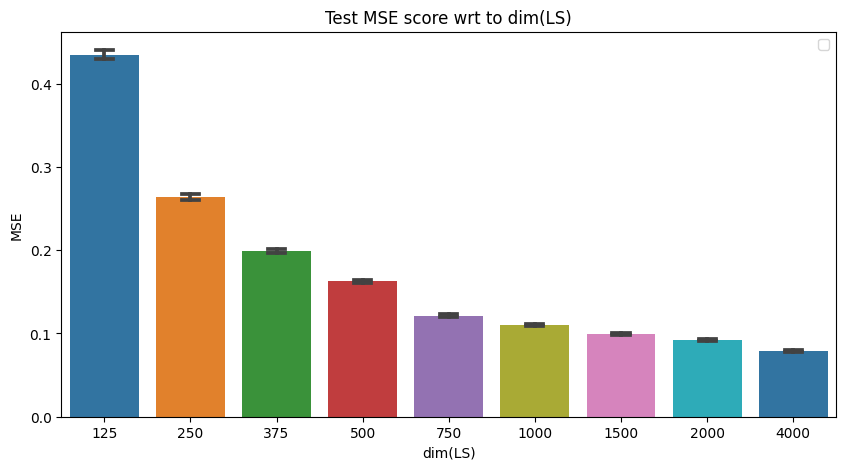}
        \caption*{(c)}
    \end{minipage}
    \caption{Effect of LSTEEG's hyperparameters on reconstruction error, averaged over all the epochs from the test partition. (a) shows the effect that different combinations of $N_o$ and $N_i$ have on MSE. (b) shows the effect on MSE of $N_o$ changes when fixing $N_i=50$. (c) shows the effect of changing $N_{LS}$, with $N_o=50$, $N_i=25$.}
    \label{fig:example}
\end{figure}

\FloatBarrier 
\clearpage 

\subsection{Temporal Activation of Latent Space Dimensions}
Firstly, we aim to investigate the temporal features of an EEG epoch encoded in the LS dimensions. Thus, we explore how each dimension in the Latent Space is "activated" by the contents of the EEG epochs. In other words, when we forward an epoch ($x_e$) through our LSTEEG's encoder, we obtain a projection a in the LS, $f_E(x_e) \in \mathbb{R}^{N_{LS}}$. We try to understand the characteristics of an EEG epoch that increase or decrease the value at each dimension $j$ in the LS, $f_E^j(x_e)$. We therefore compute the “Temporal Activation” of dimension $j$, $\alpha^j$ through:

\begin{equation}
\alpha^j = \sum_{e=1}^{N_e} x_e f_E^j(x_e)   
\end{equation}

where $N_e$ is the total number of epochs in the dataset.

We illustrate our motivation for this approach with a simplified example. Let us say that our LSTEEG has learned to encode alpha oscillations in dimension $j$. We would expect that when the input epoch shows clear alpha oscillations, the output $f_E^j(x_e)$ will be large, and vice-versa. Thus, those epochs with strong alpha oscillations will have larger weight when computing $\alpha_j$.

For visualization, we will choose those dimensions with the largest activations, that is, with the largest $\sum_{e=1}^(N_e)|f_E^j (x_e)|$.

However, since we do not have time-locked activity, we might incur into destructive interference when averaging over the entire dataset.

\FloatBarrier 
\clearpage 
\subsection{Artifact Correction Supplementary Figures}

\subsubsection{Comparison of $X_{br}$ Figures}

\begin{figure}[ht]
    \centering
    \begin{subfigure}[b]{0.495\textwidth}
        \includegraphics[width=\textwidth]{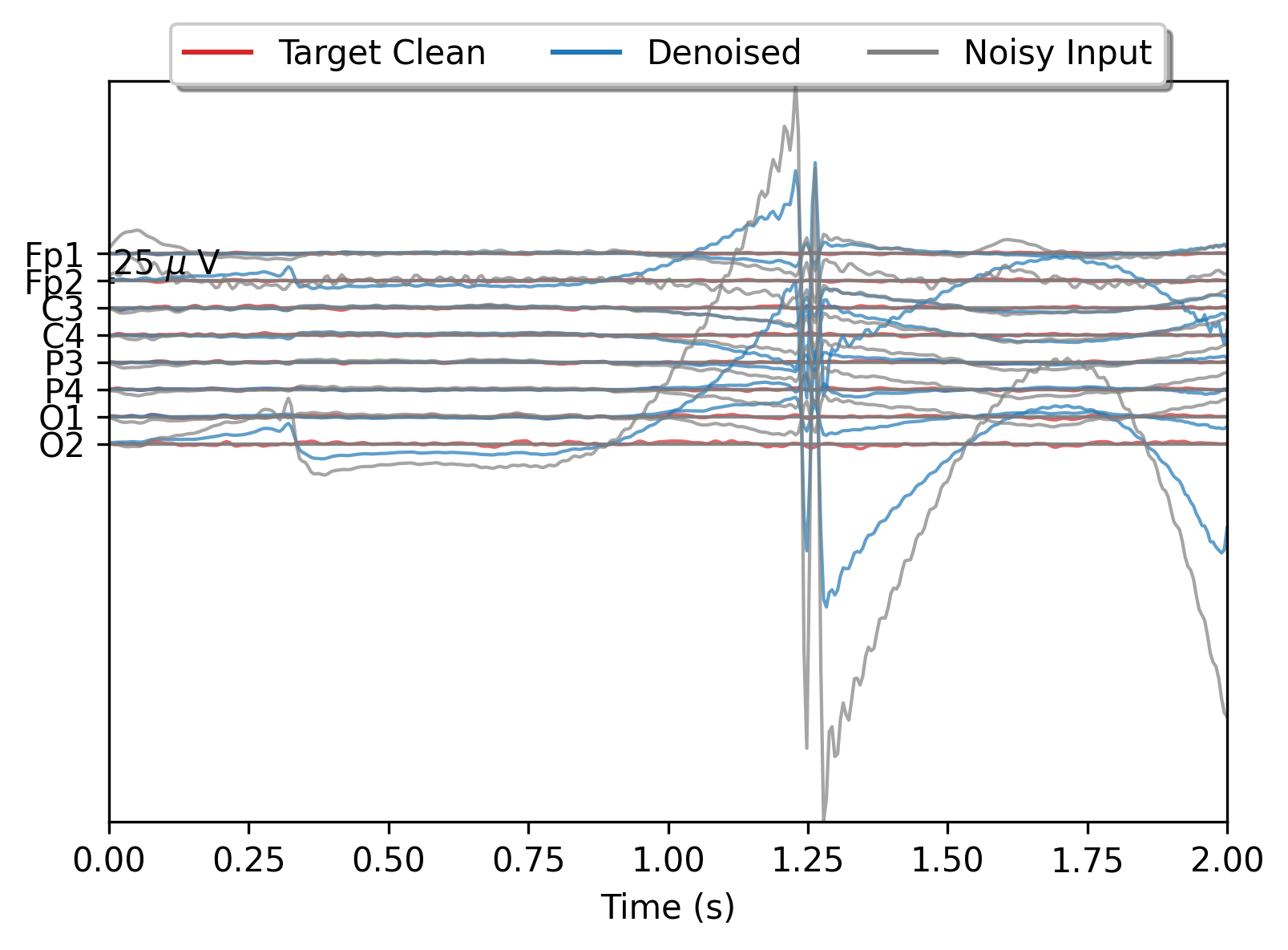}
        \caption{CLEEGN}
        \label{fig:Br_large_cleegn}
    \end{subfigure}
    \hfill 
    \begin{subfigure}[b]{0.495\textwidth}
        \includegraphics[width=\textwidth]{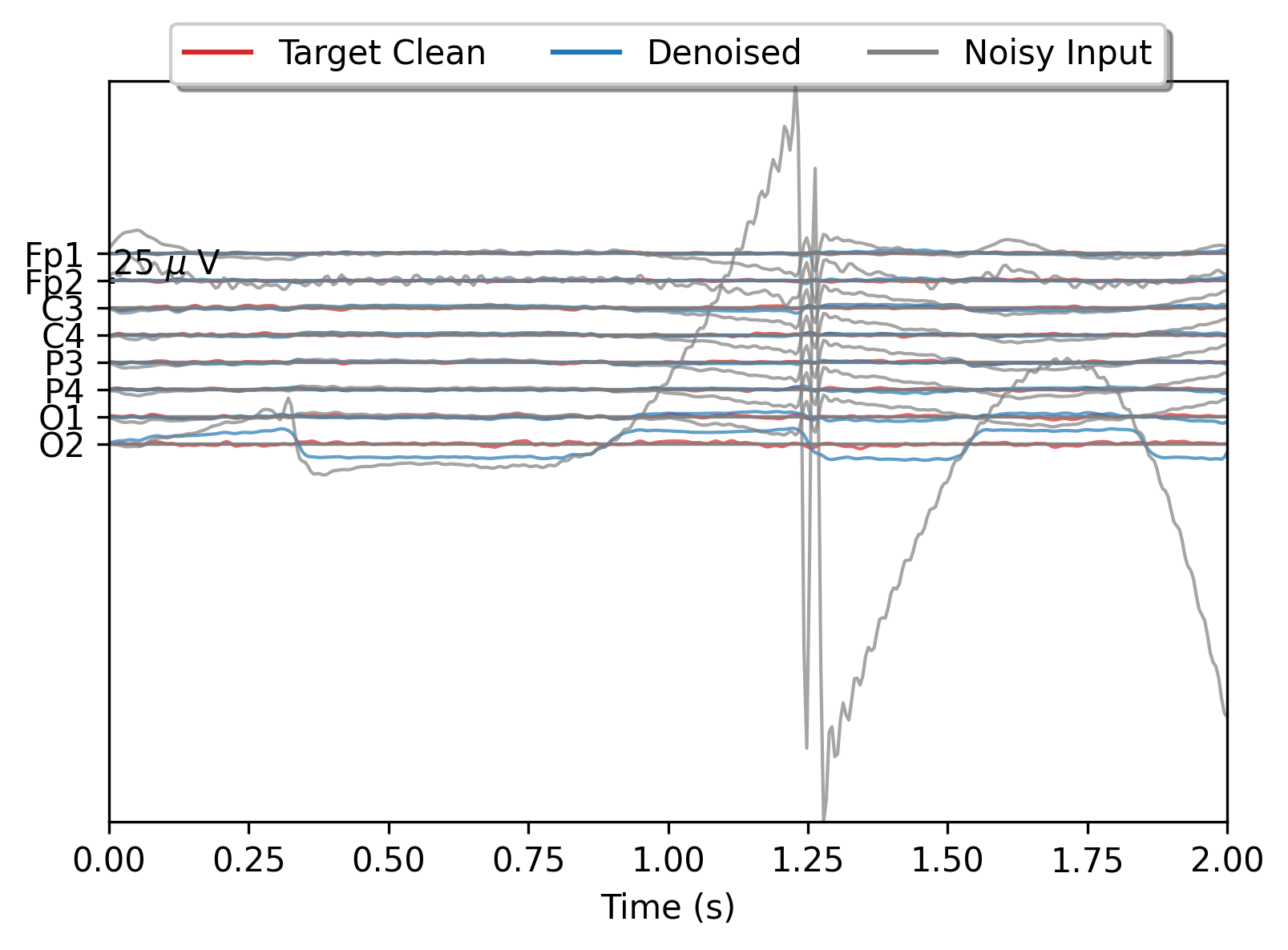}
        \caption{UNET}
        \label{fig:Br_large_unet}
    \end{subfigure}

    \begin{subfigure}[b]{0.495\textwidth}
        \includegraphics[width=\textwidth]{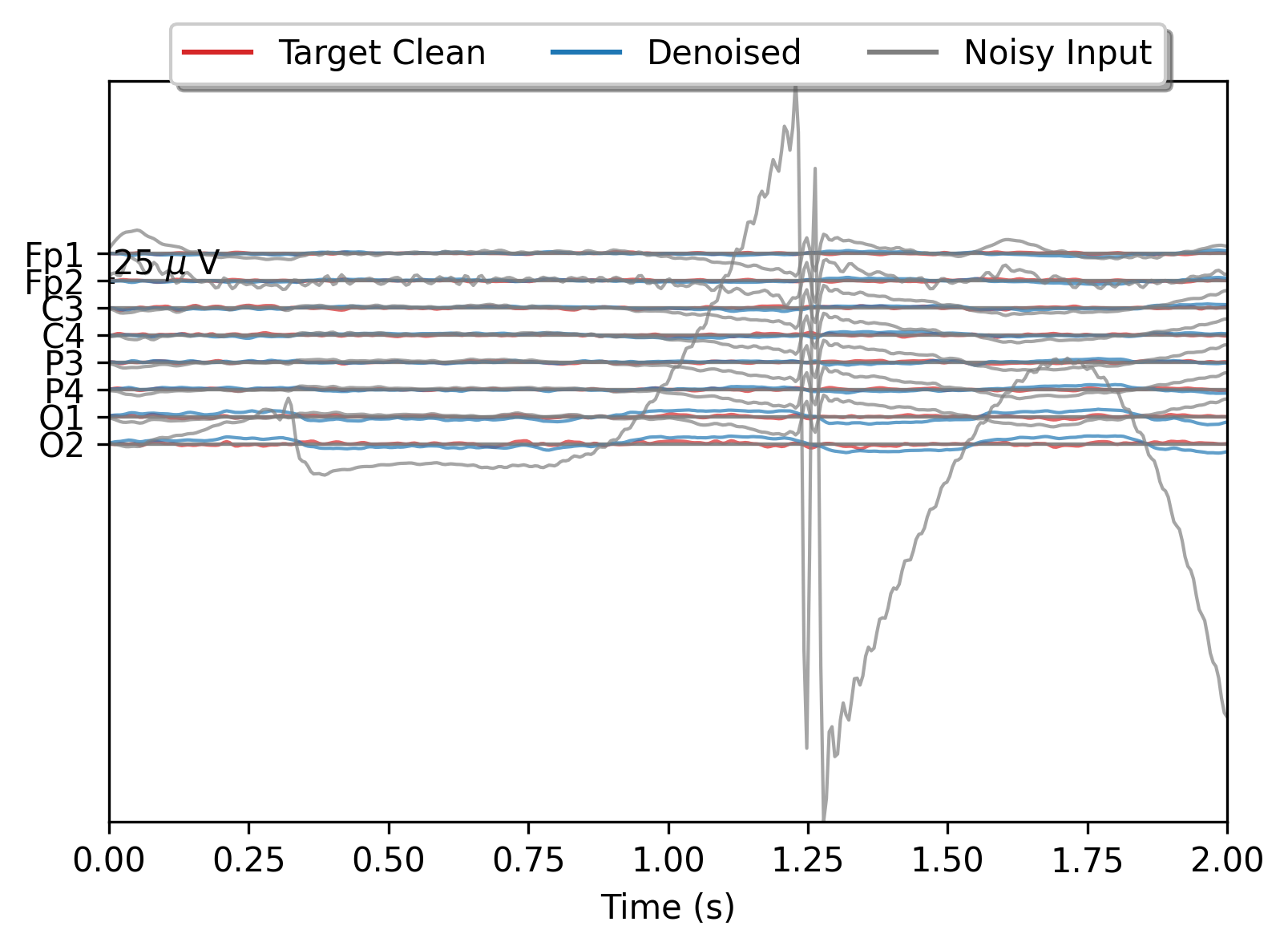}
        \caption{LSTEEG ($N_{LS}=500$)}
        \label{fig:Br_large_staraeeg500}
    \end{subfigure}
    \hfill 
    \begin{subfigure}[b]{0.495\textwidth}
        \includegraphics[width=\textwidth]{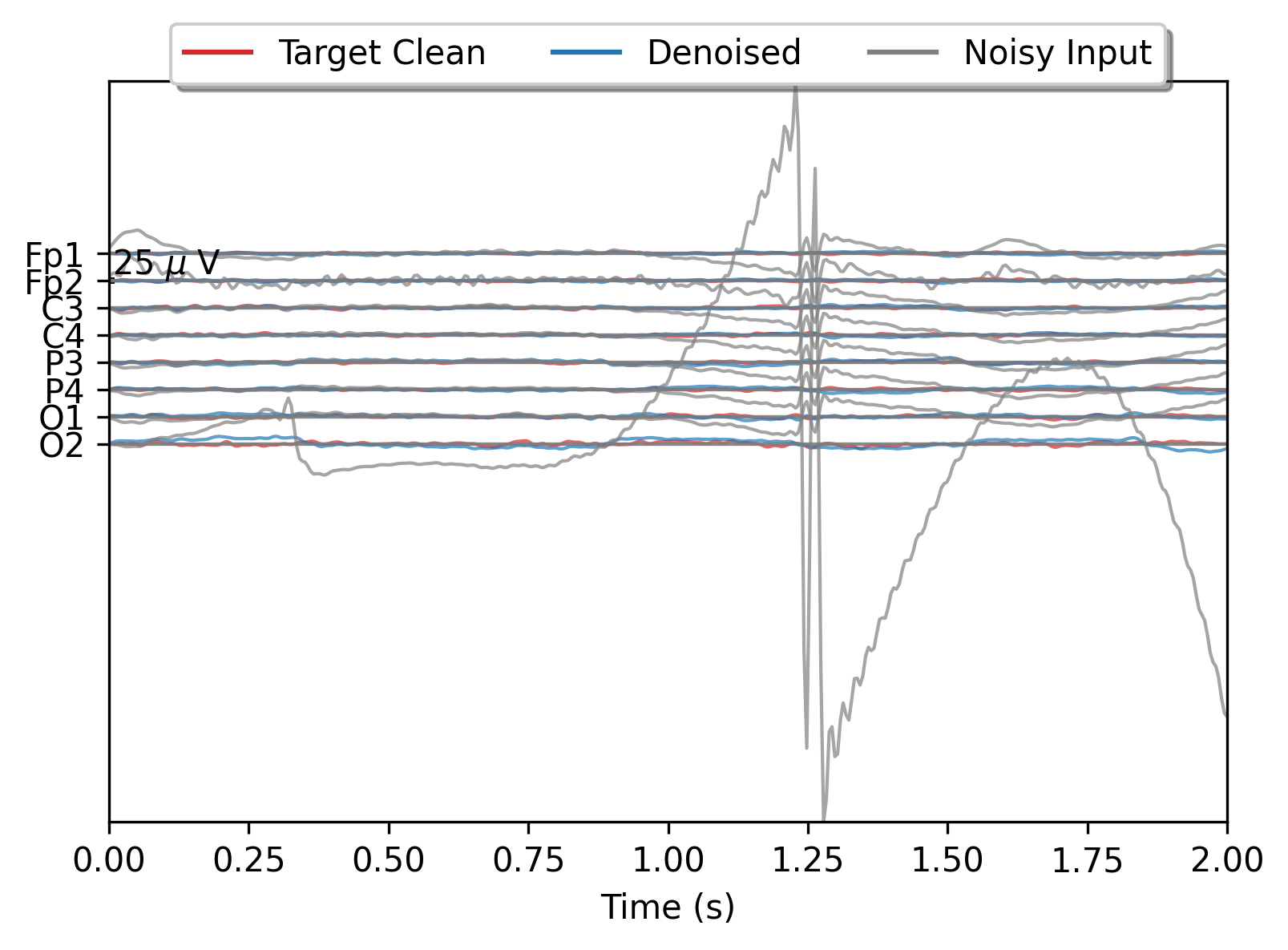}
        \caption{LSTEEG ($N_{LS}=2000$)}
        \label{fig:Br_large_staraeeg2000}
    \end{subfigure}

    \caption{Reconstruction Comparison for a large amplitude EEG artifact in the testing set. Shown networks trained with $\mathbf{X_{Br}}$. The Red line is the cleaned target EEG epoch; the Grey line is the input EEG epoch, containing the large amplitude artifact; the Blue line is the output of each network. While CLEEGN produces high amplitude outputs, unable to correct the artifact, both UNET and the two configurations of LSTEEG are able to remove the artifact from the EEG signal.}
    \label{fig:Br_large}
\end{figure}

\begin{figure}[ht]
    \centering
    \begin{subfigure}[b]{0.495\textwidth}
        \includegraphics[width=\textwidth]{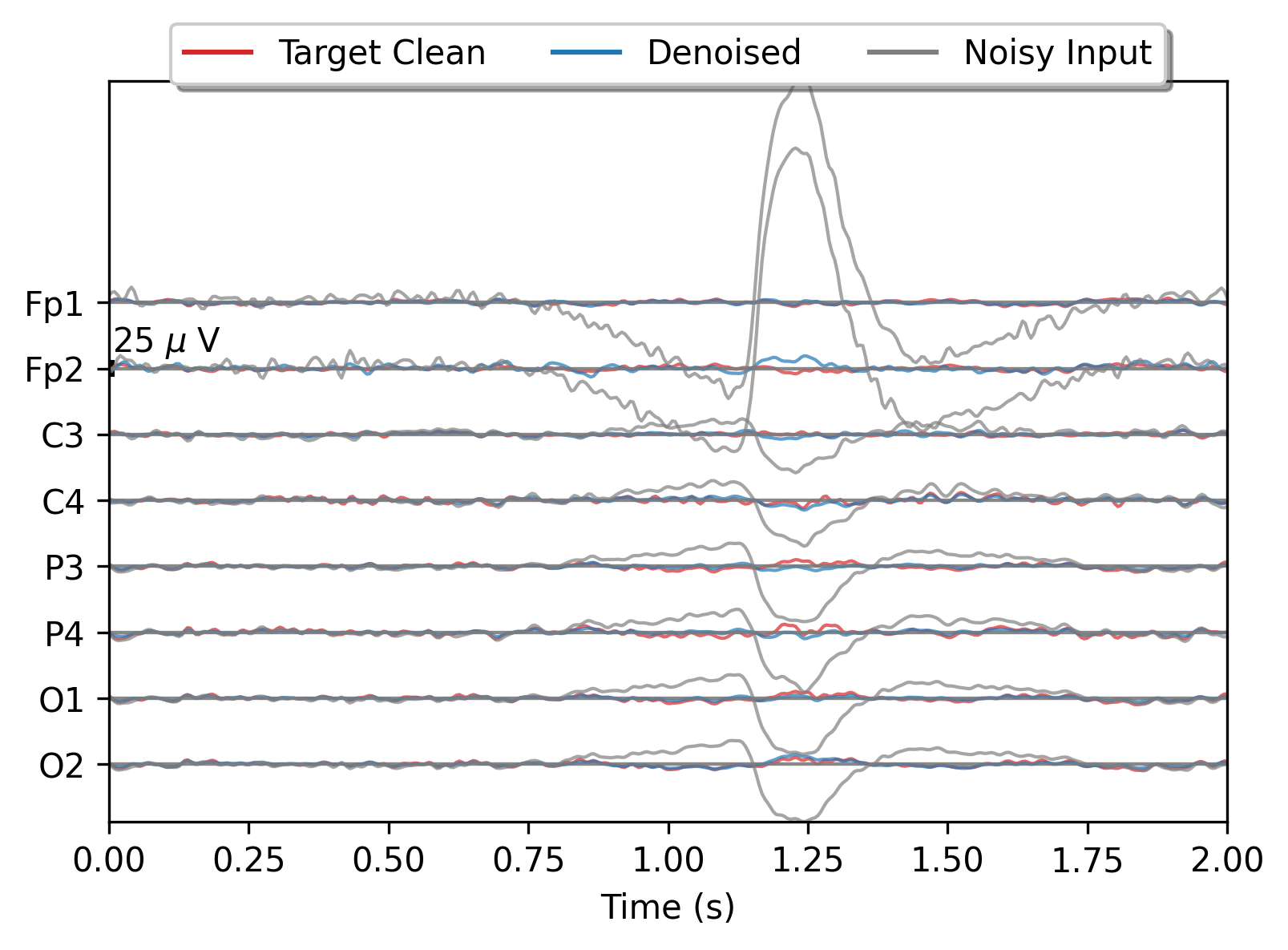}
        \caption{CLEEGN}
        \label{fig:Br_eye_cleegn}
    \end{subfigure}
    \hfill 
    \begin{subfigure}[b]{0.495\textwidth}
        \includegraphics[width=\textwidth]{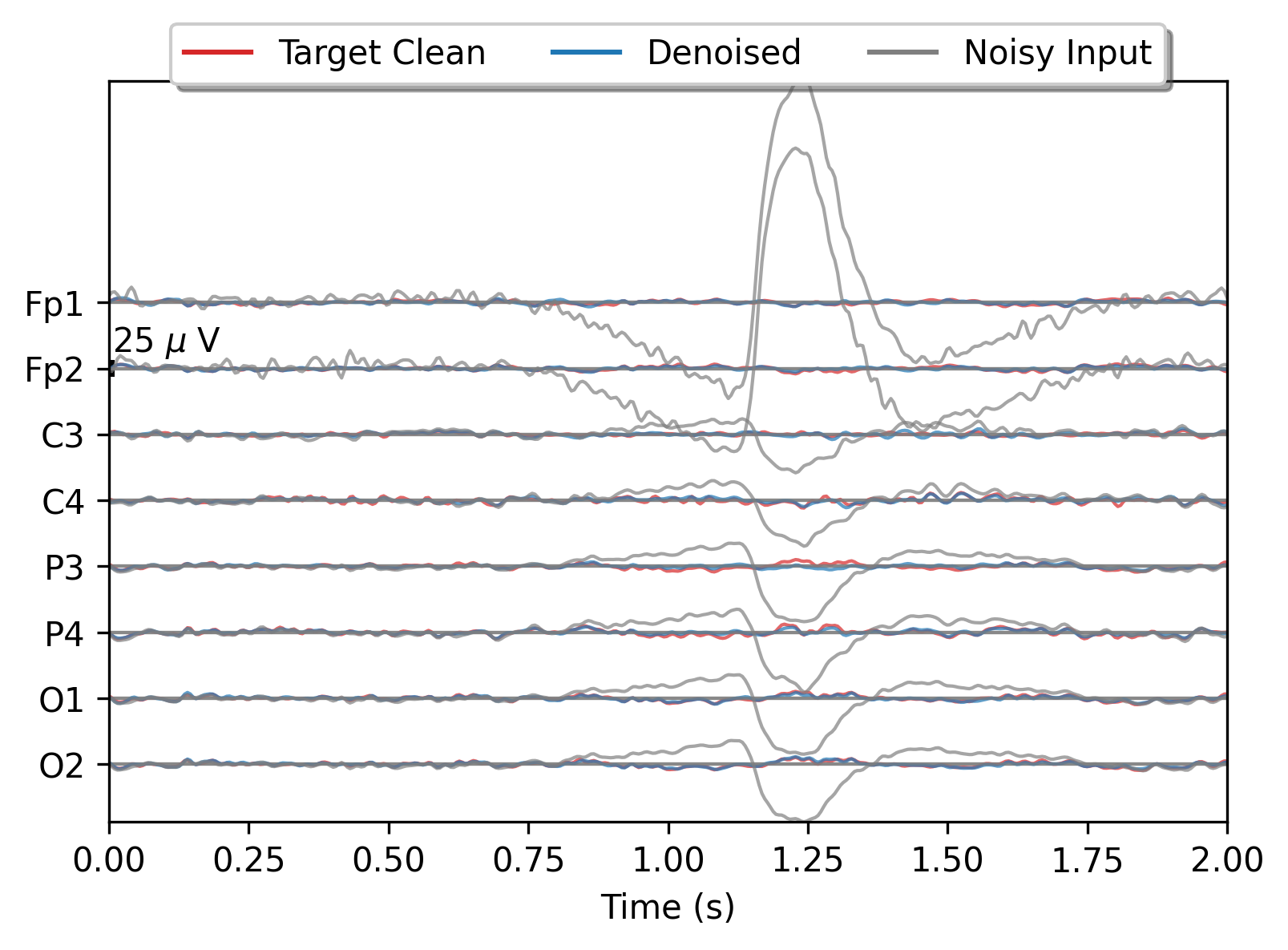}
        \caption{UNET}
        \label{fig:Br_eye_unet}
    \end{subfigure}

    \begin{subfigure}[b]{0.495\textwidth}
        \includegraphics[width=\textwidth]{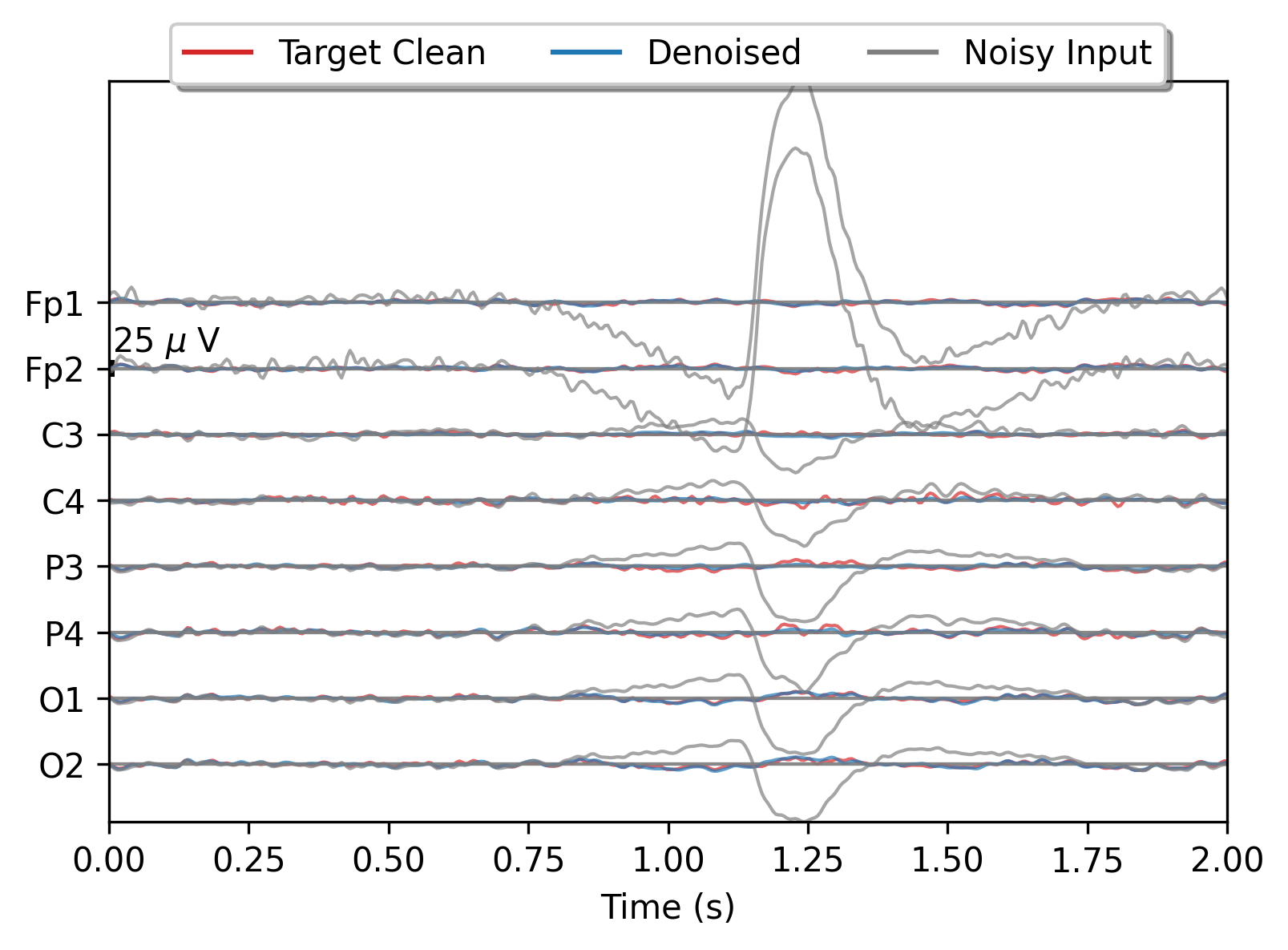}
        \caption{LSTEEG ($N_{LS}=500$)}
        \label{fig:Br_eye_staraeeg500}
    \end{subfigure}
    \hfill 
    \begin{subfigure}[b]{0.495\textwidth}
        \includegraphics[width=\textwidth]{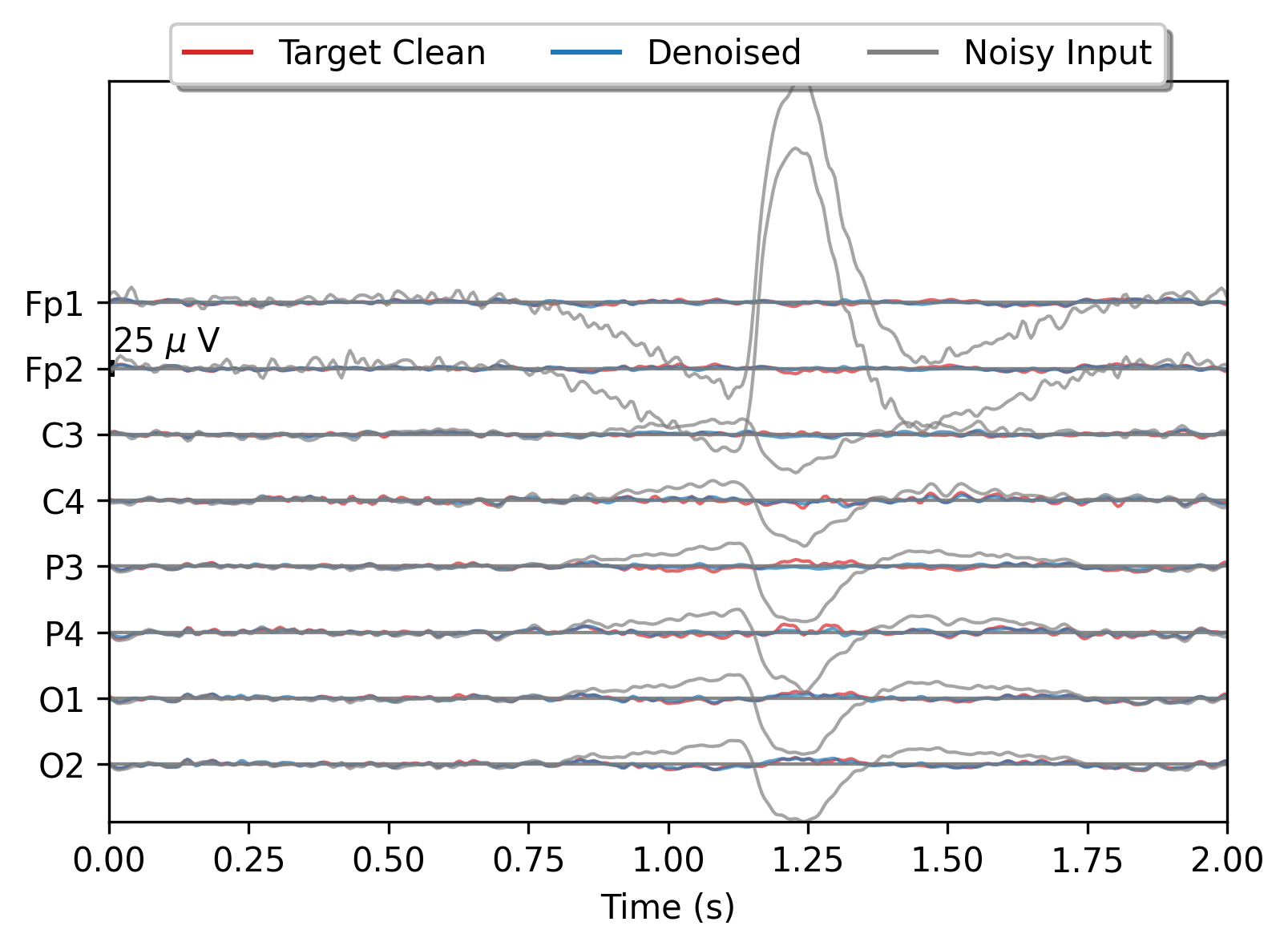}
        \caption{LSTEEG ($N_{LS}=2000$)}
        \label{fig:Br_eye_staraeeg2000}
    \end{subfigure}

    \caption{Reconstruction Comparison for a typical ocular artifact in the test set. Shown networks trained with $\mathbf{X_{Br}}$. The Red line is the cleaned target EEG epoch; the Grey line is the input EEG epoch, containing the large amplitude artifact; the Blue line is the output of each network. While CLEEGN produces high amplitude outputs, unable to correct the artifact, both UNET and the two configurations of LSTEEG are able to remove the artifact from the EEG signal.}
    \label{fig:Br_eye}
\end{figure}

\FloatBarrier 
\clearpage 

\subsubsection{Comparison of $X_{ar}$ Figures}

\begin{figure}[ht]
    \centering
    \begin{subfigure}[b]{0.495\textwidth}
        \includegraphics[width=\textwidth]{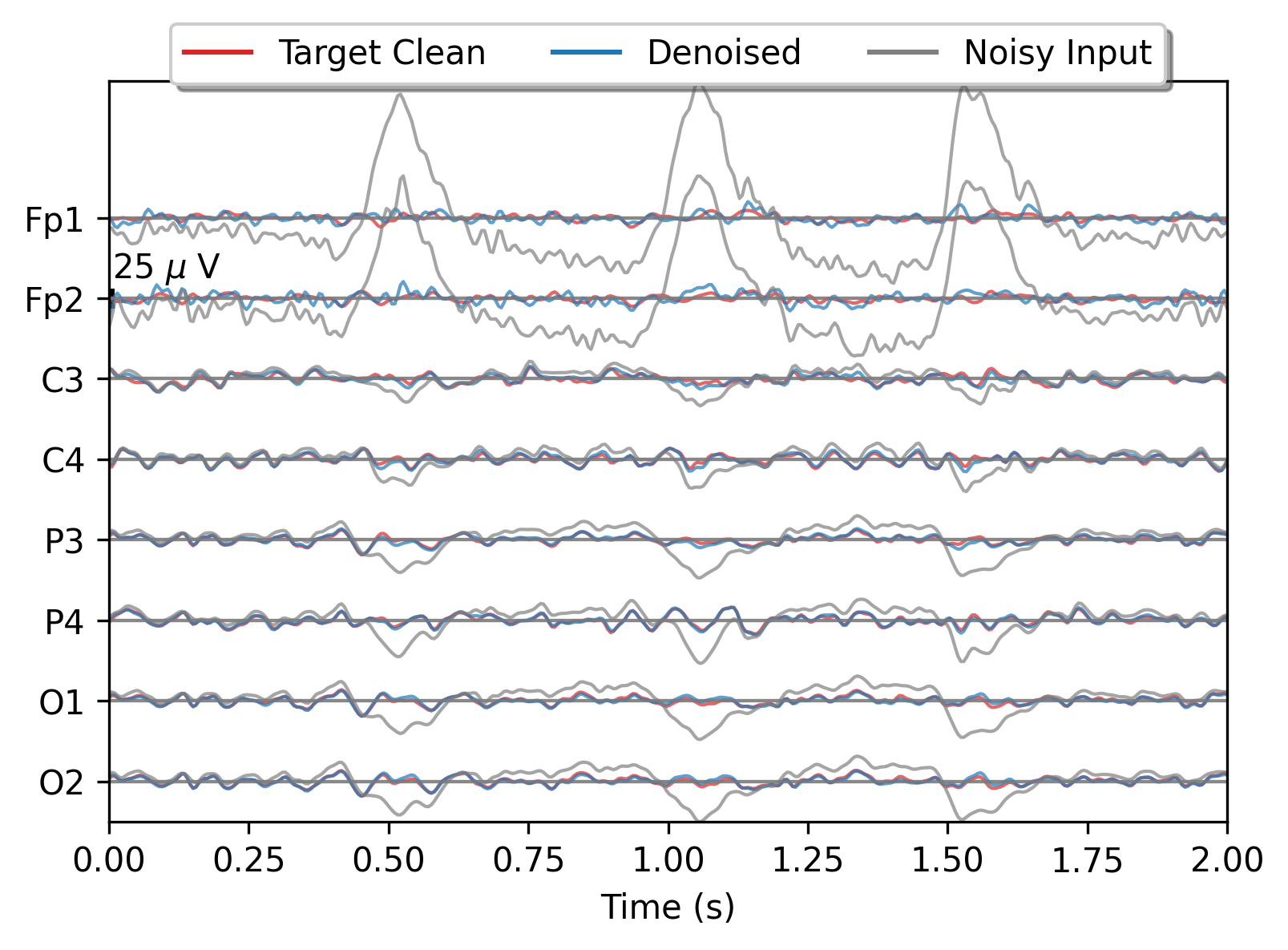}
        \caption{CLEEGN}
        \label{fig:AR_eye_cleegn}
    \end{subfigure}
    \hfill 
    \begin{subfigure}[b]{0.495\textwidth}
        \includegraphics[width=\textwidth]{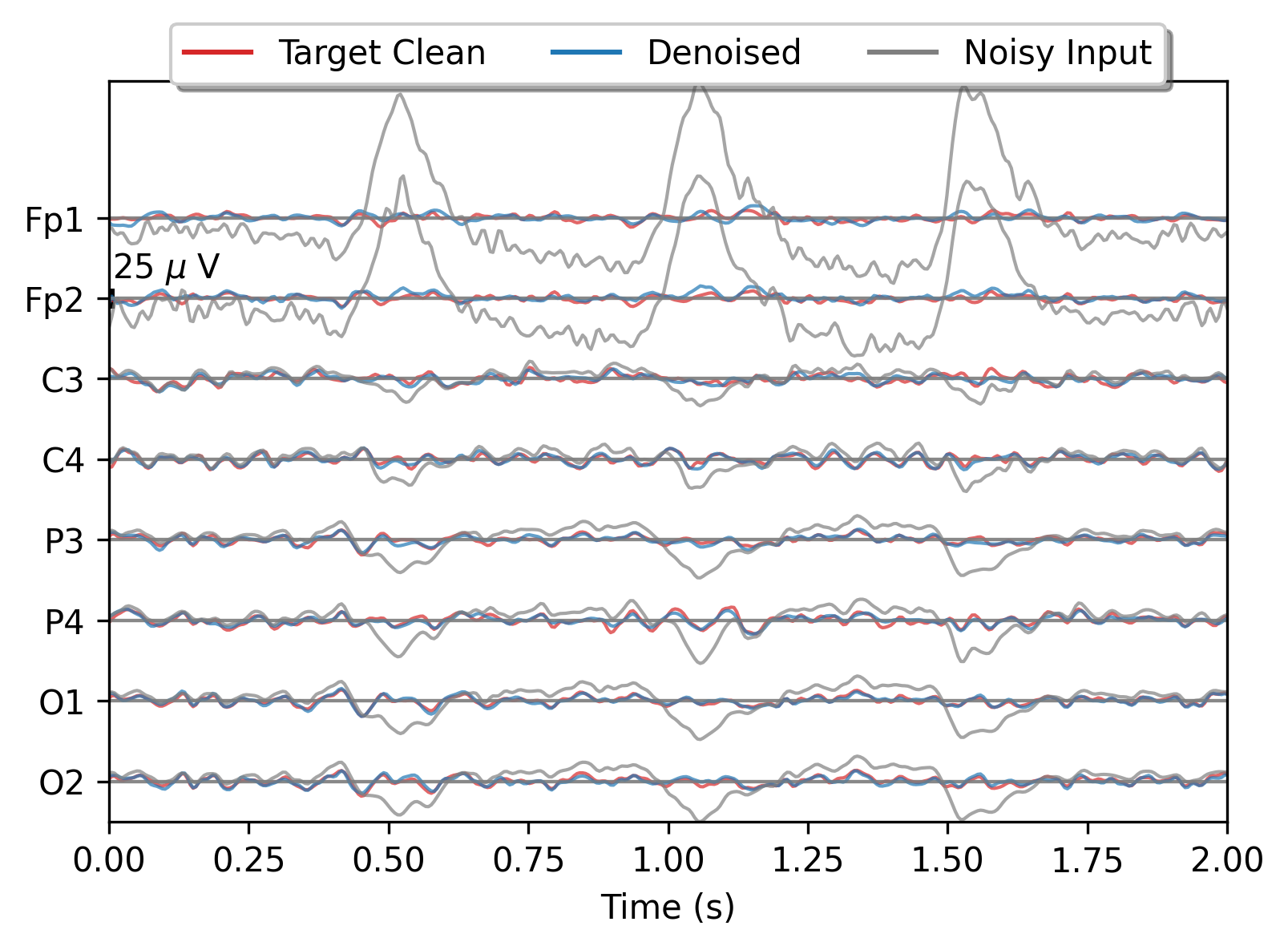}
        \caption{UNET}
        \label{fig:AR_eye_unet}
    \end{subfigure}

    \begin{subfigure}[b]{0.495\textwidth}
        \includegraphics[width=\textwidth]{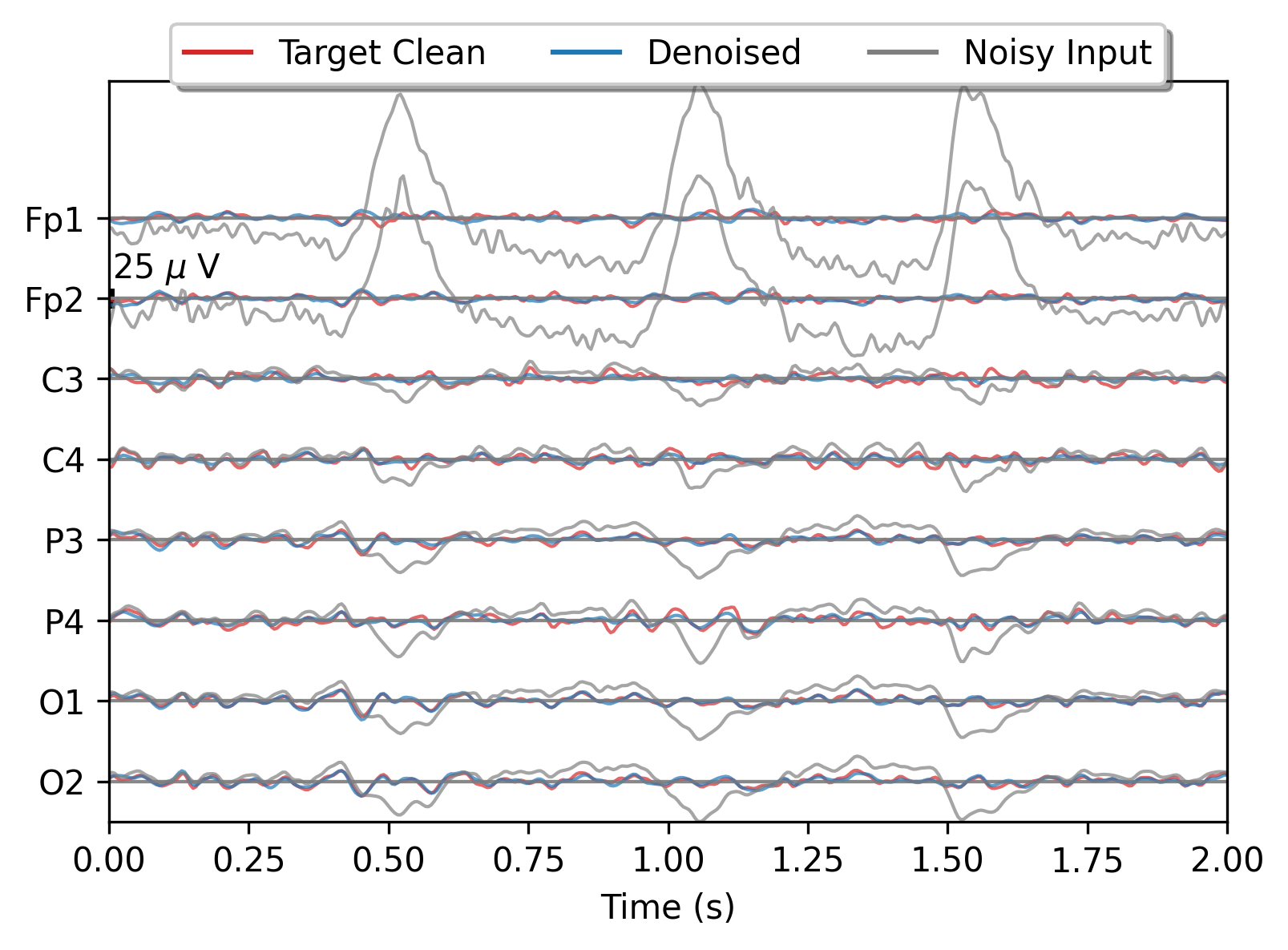}
        \caption{LSTEEG ($N_{LS}=500$)}
        \label{fig:AR_eye_staraeeg500}
    \end{subfigure}
    \hfill 
    \begin{subfigure}[b]{0.495\textwidth}
        \includegraphics[width=\textwidth]{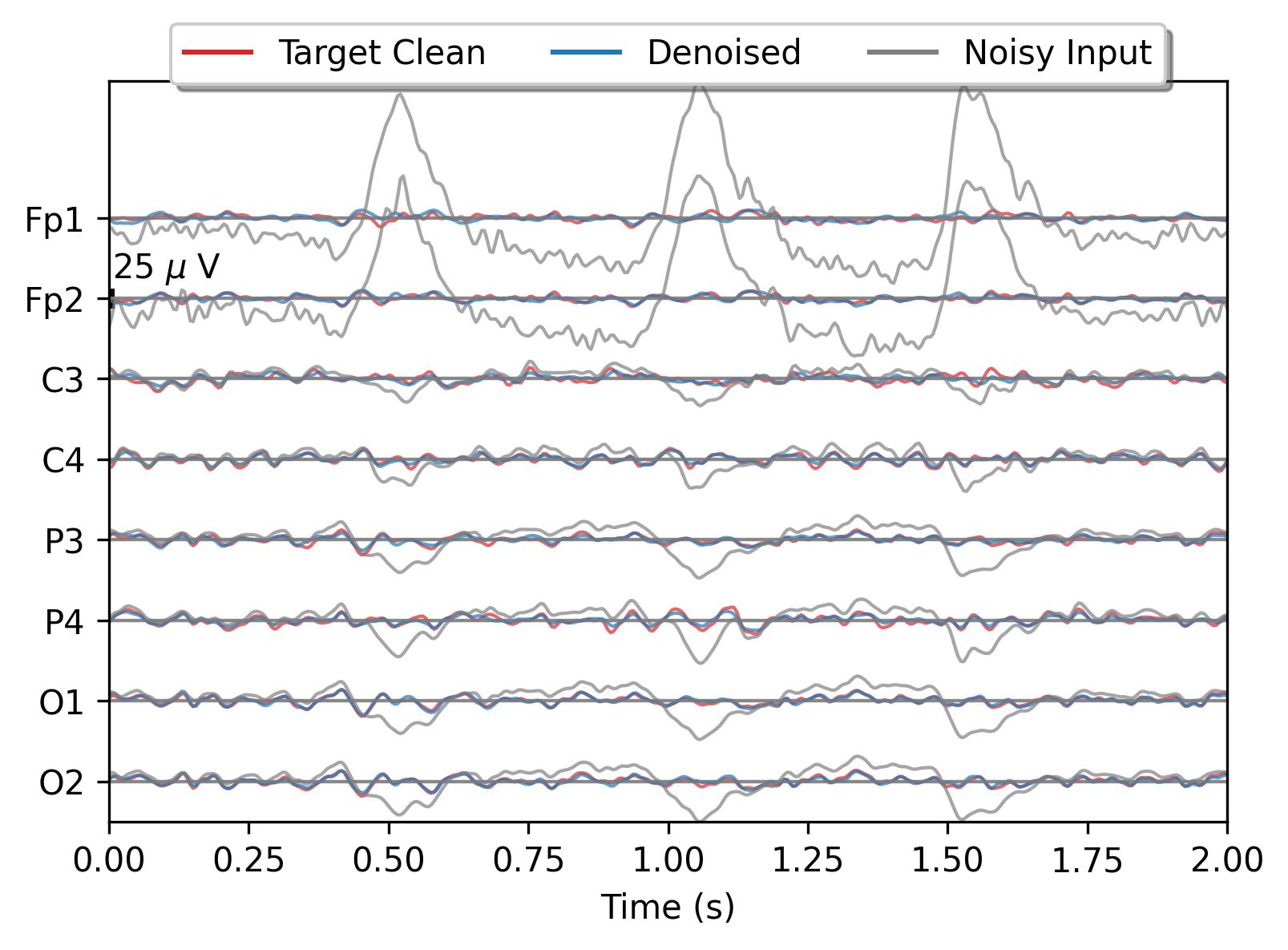}
        \caption{LSTEEG ($N_{LS}=2000$)}
        \label{fig:AR_eye_staraeeg2000}
    \end{subfigure}

    \caption{Reconstruction Comparison for a typical ocular artifact in the testing set. Shown networks trained with $\mathbf{X_{Ar}}$. The Red line is the cleaned target EEG epoch; the Grey line is the input EEG epoch, containing the large amplitude artifact; the Blue line is the output of each network. While CLEEGN produces high amplitude outputs, unable to correct the artifact, both UNET and the two configurations of LSTEEG are able to remove the artifact from the EEG signal.}
    \label{fig:AR_eye}
\end{figure}

\FloatBarrier 
\clearpage 

\subsection{Comparison of Spectral Attenuation}\label{sec:supp_attenuation}

\begin{figure}[ht]
    \centering
    \begin{subfigure}[b]{\textwidth}
        \centering
        \includegraphics[width=0.9\textwidth]{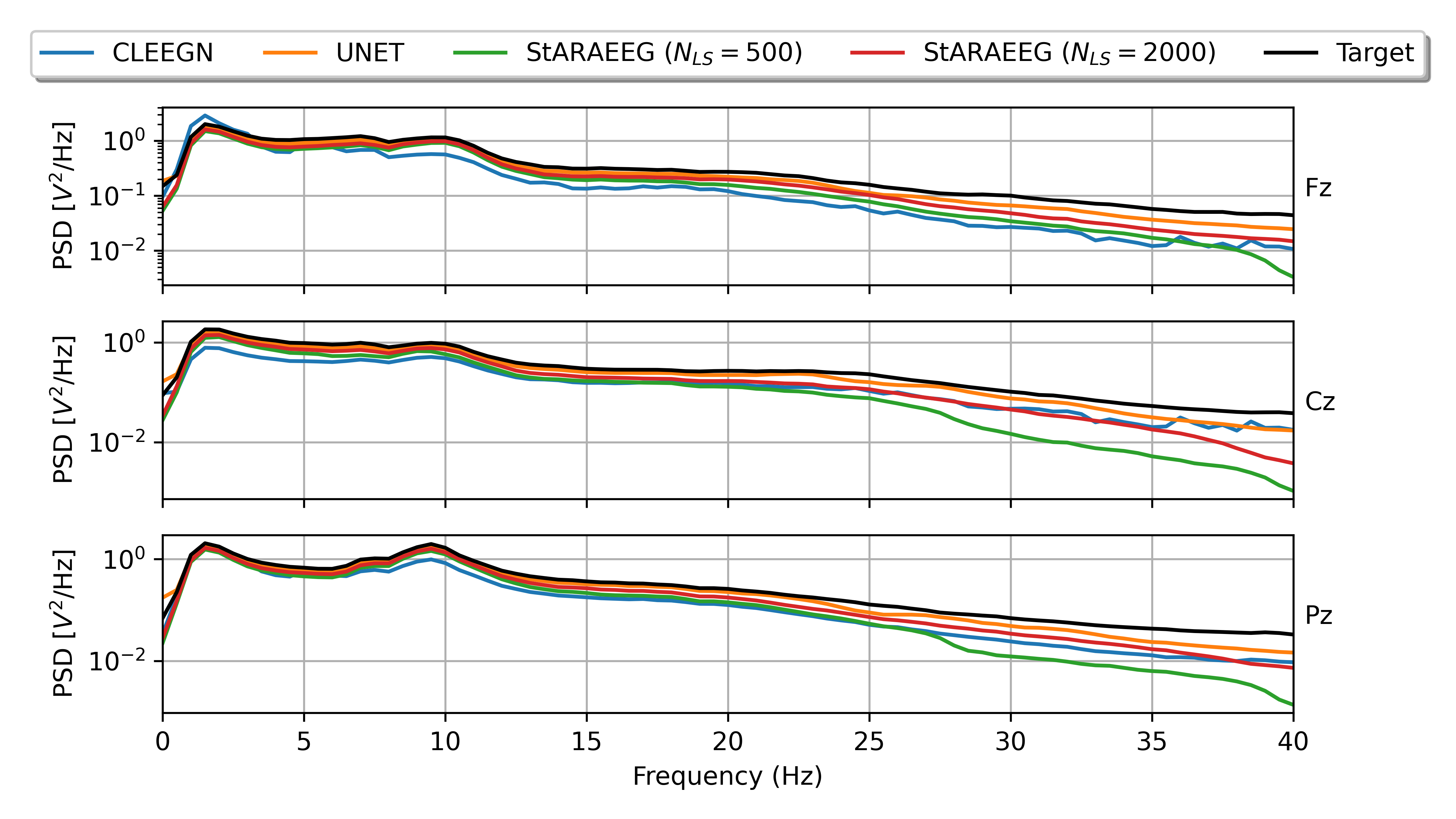}
        \caption{$\mathbf{X_{Br}}$}
        \label{fig:psd_br}
    \end{subfigure}
    \vfill
    \begin{subfigure}[b]{\textwidth}
        \centering
        \includegraphics[width=0.9\textwidth]{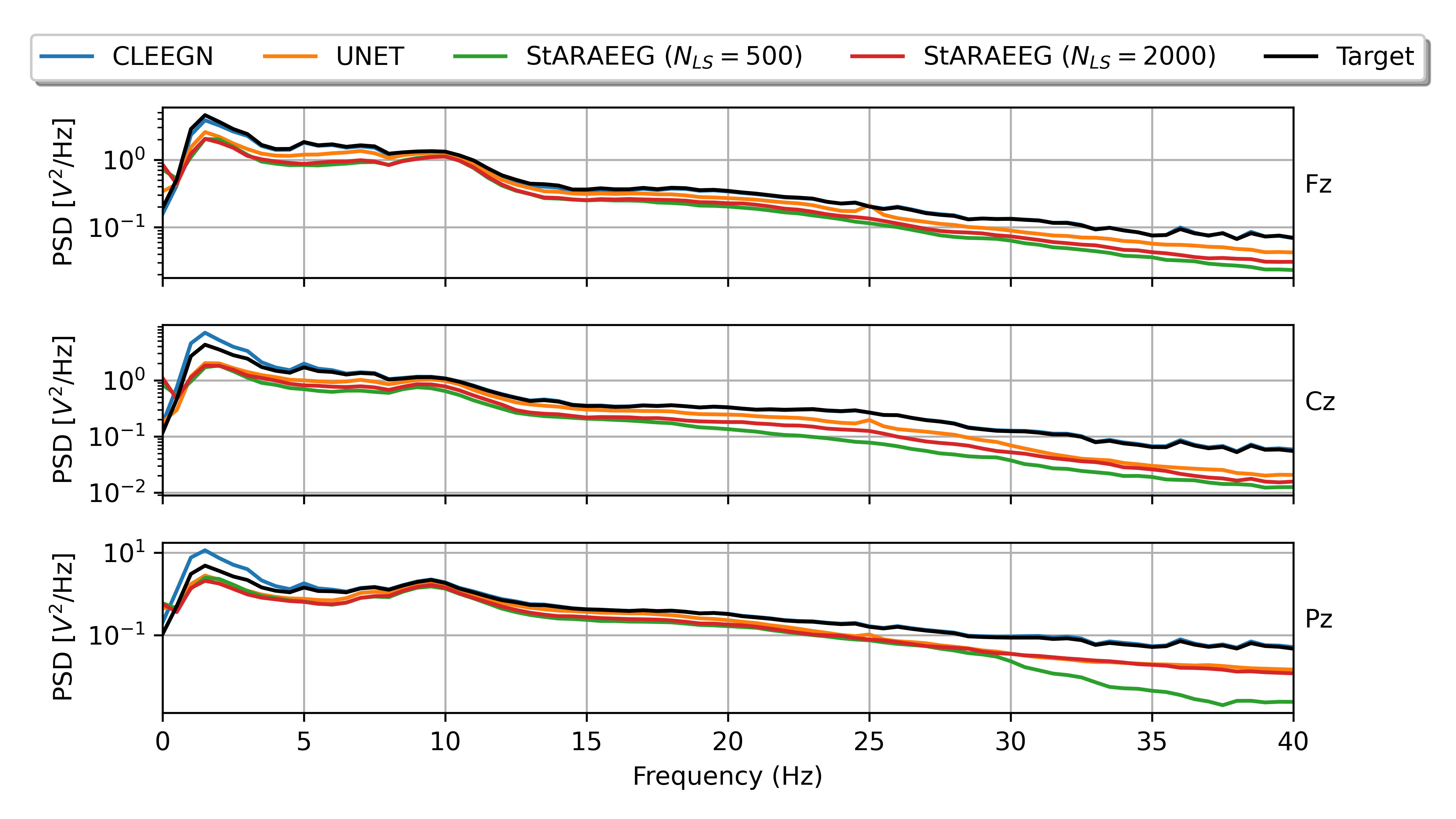}
        \caption{$\mathbf{X_{Ar}}$}
        \label{fig:psd_ar}
    \end{subfigure}
    \caption{PSD Attenuation for UNET and LSTEEG. One can note the sharp decay of LSTEEGs ($N_{LS}=500$) PSD after a certain frequency.}
    \label{fig:supp_attenuation}
\end{figure}